\documentclass{article}
\usepackage{arxiv}
\usepackage[utf8]{inputenc}
\usepackage[numbers,square]{natbib}
\usepackage{amssymb}
\usepackage{amsthm}

\usepackage{array}

\usepackage{algorithm}
\usepackage{algpseudocode}
\usepackage{amsmath}
\usepackage{subfig}
\usepackage{xurl}
\usepackage[table]{xcolor}
\usepackage{booktabs,array,amsmath,graphicx,microtype,mathptmx}
\usepackage{multirow}
\usepackage[T1]{fontenc}
\usepackage[labelfont=bf,font=small]{caption}
\renewcommand{\arraystretch}{1.05}

\algnewcommand\algorithmicforeach{\textbf{for each}}
\algdef{S}[FOR]{ForEach}[1]{\algorithmicforeach\ #1\ \algorithmicdo}

\usepackage{booktabs,graphicx,xcolor,colortbl,rotating}

\usepackage[hidelinks]{hyperref}

\begin{document}

\title{PUMA: Learning a Mutation-Aware Vocabulary of Protein Units}
\date{}

\author{%
  \parbox{\dimexpr\textwidth-2\tabcolsep\relax}{\centering
  \begin{tabular*}{\linewidth}{@{\extracolsep{\fill}}ccc@{}}
    Burak Suyunu$^{1,\dagger}$ & Özdeniz Dolu$^{1,\dagger}$ & Ibukunoluwa Abigail Olaosebikan$^{2}$
  \end{tabular*}\\[12pt]
  \begin{tabular*}{0.62\linewidth}{@{\extracolsep{\fill}}cc@{}}
    Hacer Karatas Bristow$^{2,3,4}$ & Arzucan Özgür$^{1,\ast}$
  \end{tabular*}\\[16pt]
  {\mdseries\small
  \makebox[\linewidth][c]{$^{1}$Department of Computer Engineering, Boğaziçi University, Bebek, 34342, İstanbul, Türkiye}\\
  \makebox[\linewidth][c]{$^{2}$C. Eugene Bennett Department of Chemistry, West Virginia University, Morgantown, 26505, West Virginia, United States}\\
  \makebox[\linewidth][c]{$^{3}$Department of Pharmaceutical Chemistry, Faculty of Pharmacy, Istanbul Medipol University, Beykoz, 34815, İstanbul, Türkiye}\\
  \makebox[\linewidth][c]{$^{4}$Research Institute for Health Sciences and Technologies (SABITA), Istanbul Medipol University, Beykoz, 34815, İstanbul, Türkiye}\\[8pt]
  $^{\dagger}$These authors contributed equally to this work. \\
  $^{\ast}$Corresponding author: \href{mailto:arzucan.ozgur@bogazici.edu.tr}{arzucan.ozgur@bogazici.edu.tr}}%
  }%
}

\maketitle

\begin{abstract}
Modeling protein sequences as a language has made language models a powerful tool in computational biology, yet the language itself remains poorly understood. A key step toward understanding it is identifying its constituent units. In natural languages, morphemes can occur in multiple forms; similarly, in proteins, mutations can give rise to variations of a unit that persist through evolution, forming families of related units.
We introduce PUMA (Protein Units via Mutation-Aware Merging), an algorithm that learns protein units from sequence and explores their mutational variants using substitution matrices, forming a genealogy of unit families. Our results show that mutations remaining within a PUMA family are more often benign than the substitution matrix alone predicts, and that PUMA genealogy improves molecular function representations compared to treating units independently. A case study of a unit family demonstrates relatedness beyond homology. PUMA achieves competitive performance on downstream tasks when used as a protein language model tokenizer. Moreover, collapsing units into families results in a smaller embedding table and faster training. Together, these results support PUMA as a biologically grounded protein vocabulary that organizes protein units into plausible families of mutational variants.
The source code is available at \url{https://github.com/boun-tabi-lifelu/PUMA}.
\end{abstract}

\section{Introduction}
\label{sec:introduction}


The amino acid sequences of proteins have long been viewed as text written in a language~\citep{bralley1996introduction}, and protein language models now process these sequences as such~\citep{hayes2024simulating}. Extending this analogy, we seek to identify units of protein sequence that play a role analogous to words or morphemes in natural language: short stretches of sequence that recur across proteins and are reused as building blocks. As proteins evolve through mutation, we expect such units to occur not as single fixed strings but as families of related variants. We therefore present an algorithm designed to discover these units and their variants directly from sequence data.

The search for such units is not new. The classical concepts of domains and motifs describe high-level functional or structural regions that may be sparse or non-contiguous, and were never designed to segment an entire sequence into its smallest contiguous units of information. This gap motivated the Protein Unit (PU) paradigm, which posits more granular, contiguous blocks forming the true structural and evolutionary basis of proteins~\citep{gelly2006protein}. The line of work has since matured into hierarchical decompositions of 3D structures into PUs that can be reassembled into domains~\citep{cretin2022sword2} and, more recently, into unit discovery driven by functional supervision~\citep{uludougan2026puffin}; in parallel, structure-derived alphabets have been used to build discrete vocabularies for protein language models~\citep{su2024saprot}. What unites these efforts is their dependence on the folded structure, whether experimental or predicted. Proteins are indeed three-dimensional biomolecules, but as Anfinsen emphasized, in a given environment the native structure is determined by the amino acid sequence~\citep{anfinsen1973principles}. A parallel question therefore remains open: whether fundamental units can be discovered directly from the 1D sequence, by observing the patterns of variation and relatedness that evolution has left in it.
We pursue this sequence-only route, which applies to any sequence, even one without a reliable structure.

\begin{figure*}[!t]
    \centering
    \includegraphics[width=\textwidth]{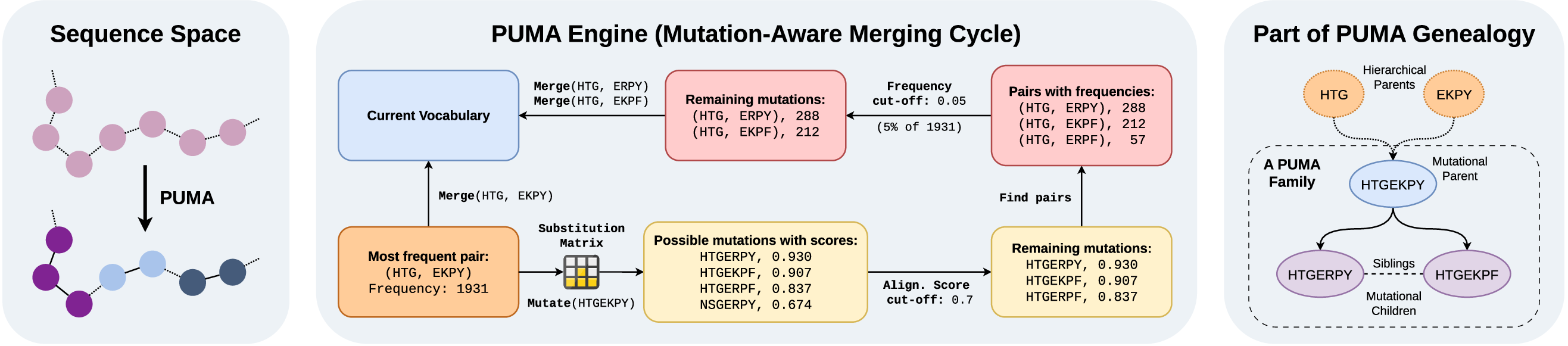}
    \caption{Overview of the PUMA framework.
    \textbf{Sequence Space:} PUMA operates directly on 1D amino acid sequences, segmenting them into contiguous Protein Units.
    \textbf{PUMA Engine:} An iterative cycle integrating frequency and molecular evolution — the most frequent pair of units is merged into a parent; substitutions of that parent are simulated from a substitution matrix; candidates are filtered by an alignment-score cut-off and a frequency cut-off relative to the parent; survivors enter the vocabulary.
    \textbf{PUMA Genealogy} maps the hierarchy and mutational divergence of units. Hierarchical parents merge to form a new unit. A PUMA family consists of a central Mutational Parent and its resulting Mutational Children (appearing as Siblings to one another). This genealogy ensures that units and their plausible mutational variants are recognized as related entities rather than distinct fragments.}
    \label{fig:PUMA-overiew}
\end{figure*}

Like natural languages, which change through dialect and usage, protein sequences are shaped by molecular evolution. A fundamental unit of protein language is therefore not a static string but a family of mutational variants that preserve function despite sequence divergence — in linguistic terms, a morpheme. A morpheme is the smallest meaningful unit of a language, and the same morpheme may surface in more than one written form: the words \textit{leaf} and \textit{leaves} share the morpheme \textit{leaf} but spell it differently. What we seek in protein sequences are morphemes in this sense, each with one commonly occurring surface form and a set of related variants realizing the same meaning, recognized as related units rather than as distinct entities. This has been a conceptual limitation for tokenization methods adapted from natural language processing (NLP), such as Byte-Pair Encoding (BPE)~\citep{sennrich2016neural}, which are powerful but guided primarily by data frequency and largely void of biological domain knowledge~\citep{dotan2024effect, suyunu2024linguistic}. Where biology has been injected, it has typically taken the form of reducing the amino acid alphabet rather than organizing the vocabulary itself~\citep{rannon2026optimizing}.


We introduce PUMA (Protein Units via Mutation-Aware Merging), a data-driven, frequency-based, iterative merging algorithm that incorporates mutational relationships to discover ``morphemes'' in protein language.
The algorithm produces essentially two outputs: a vocabulary of protein units and a genealogy that organizes them into families, in which each unit and its plausible mutational variants are explicitly linked (Figure~\ref{fig:PUMA-overiew}). Furthermore, a trained PUMA vocabulary can segment any protein sequence into its units without any gaps.


We evaluate PUMA along multiple dimensions, including the functional coherence of its families using clinical variants, its impact on molecular function representation, a case study of a specific PUMA family, and its effectiveness as a protein language model tokenizer. Our results establish PUMA as a sequence-only vocabulary organized by evolutionary relationships rather than frequency alone, providing a biologically grounded lexicon for protein language.

\section{PUMA}
\label{sec:puma}

PUMA builds a structured vocabulary of protein units through an iterative discovery process (pseudocode in Algorithm~\ref{sup:algo:PUMA}) that moves beyond simple frequency-based tokenization methods such as BPE~\citep{sennrich2016neural} by simulating molecular diversification at each step.
Central to this process is the construction of a ruleset, which underlies both vocabulary construction and subsequent sequence tokenization.

Starting with a vocabulary of standard amino acids as the initial units, at each iteration the algorithm expands the vocabulary by identifying the most frequently co-occurring pair of units in the dataset and merging them into a new parent unit. The corresponding merge rule is then added to the ruleset.
Once a parent unit is formed, PUMA explores its mutational neighborhood by simulating amino acid substitutions guided by matrices such as BLOSUM~\citep{henikoff1992amino} or PAM~\citep{dayhoff197822}.
For each residue in the parent unit, candidate substitutions with non-negative matrix scores are considered.
To prevent the vocabulary from expanding into biologically implausible space, each candidate mutation must satisfy two filters:
The candidate must maintain a normalized global pairwise alignment score with the parent above a cut-off $a$ (e.g., 0.7), and it must appear in the dataset with a frequency exceeding a cut-off $f$ (e.g., 5\%) relative to the parent’s frequency.
These constraints aim to preserve the unit’s structural and chemical integrity while favoring variants successful enough to persist in nature.
Qualifying candidates are added to the vocabulary, and their associated mutational rules are added to the ruleset, explicitly linking each child unit to its parent. The specific merge rules that generate these children are also recorded to facilitate consistent parsing (i.e. tokenization) of an input sequence. This cycle repeats until the target vocabulary size is reached.
To handle computational complexity, we use a max-heap to track pair frequencies and a depth-first search algorithm with pruning to optimize exploration of the exponential mutation space. A schematic of a single iteration of PUMA is shown in the middle panel of Figure~\ref{fig:PUMA-overiew}.

The primary output of this process is a genealogy where units are organized into PUMA families. A PUMA family is defined as a group of units consisting of a mutational parent (the unit formed by merging) and its mutational children (the variants generated by simulated substitutions).
Child units that share the same mutational parent are denoted as siblings.
A PUMA family is the concrete realization of the morpheme analogy: the parent is the surface form, the children its alternative spellings.
Units whose proposed mutations all fail the filters, and residues that remain unmerged, become singletons — families of size one, for which the training data supported no variants under the given thresholds.
A trained PUMA tokenizes a new sequence by applying the stored merge rules in discovery order.


\section{Experiments and Results}
\label{sec:experiments}
The PUMA vocabularies analyzed in this section are trained on a filtered subset of human protein sequences (Taxon ID: 9606) derived from UniProtKB. To mitigate redundancy, we utilized the UniRef50 dataset~\citep{suzek2015uniref} (release 2024-06), which ensures no sequence pair shares more than 50\% identity.
For each cluster, we selected the representative sequence if it was human; otherwise, we searched the cluster for a human sequence to serve as the representative.
We excluded isoforms, sequences containing more than one non-standard amino acid, and sequences exceeding 3,000 residues. The final dataset comprises 70,453 protein sequences ($\sim$22M residues) with an average length of 314 residues.
Section~\ref{sec:plm} uses the full UniRef50 release for protein language modeling.

\subsection{Properties of PUMA Vocabularies}
\label{sec:vocab_props}

\begin{figure}[!htbp]
    \begin{center}
    \includegraphics[width=.62\textwidth]{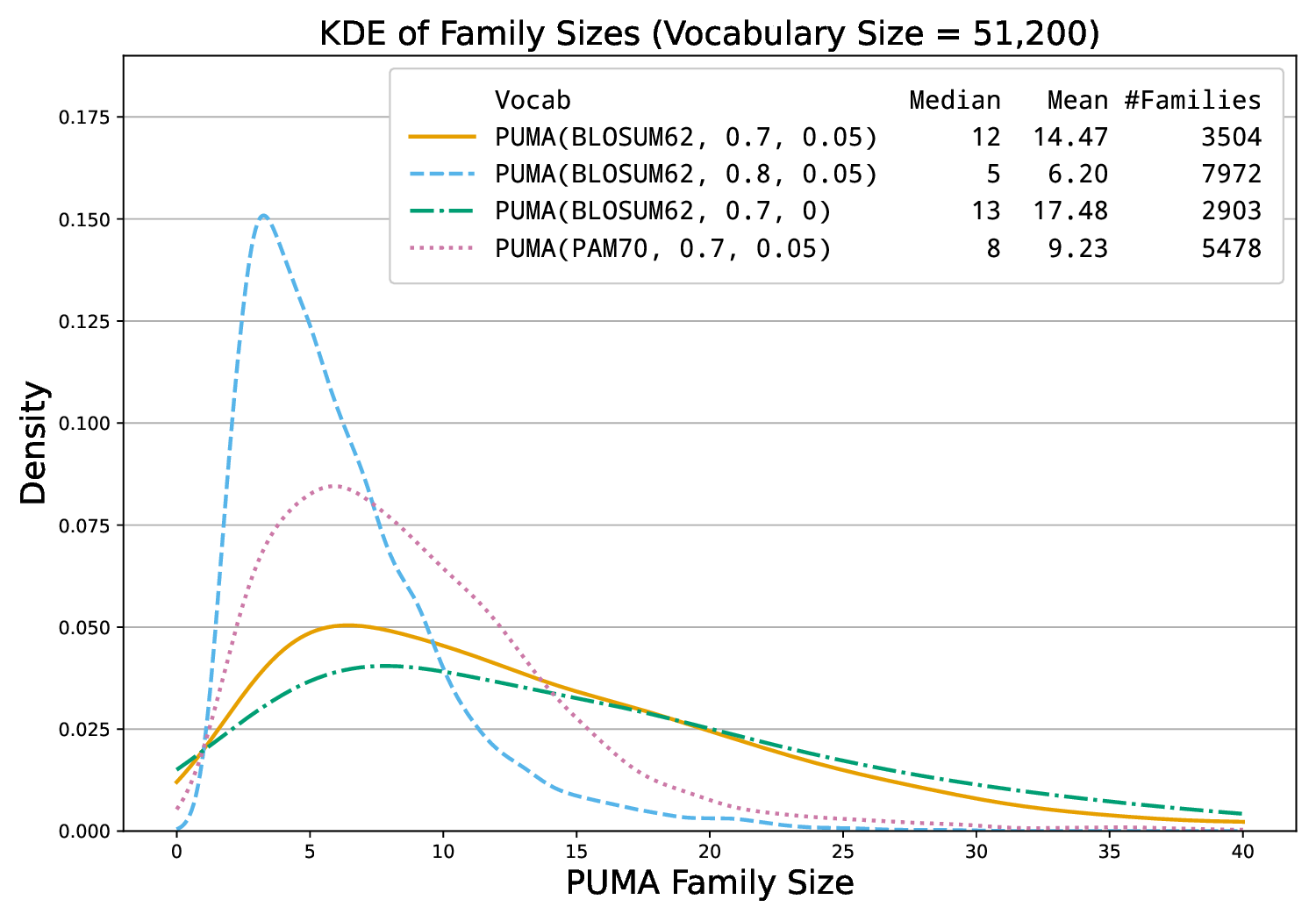}
    \caption{Distribution of PUMA family sizes at vocabulary size 51,200, for four configurations each varying one parameter off the PUMA(BLOSUM62, 0.7, 0.05) reference: substitution matrix (PAM70), alignment cut-off (0.8), and frequency cut-off (f = 0). Legend gives median and mean family size and the number of families.
    }
    \label{fig:prelim_kde}
    \end{center}
\end{figure}

Figure~\ref{fig:prelim_kde} shows the distribution of units across families for a vocabulary size of 51,200 for four configurations, each varying a single parameter relative to the PUMA(BLOSUM62, 0.7, 0.05) reference. The three parameters in a PUMA configuration correspond to the substitution matrix, alignment cut-off ($a$), and frequency cut-off ($f$), respectively.   
The results show that families account for more than 94\% of units across configurations, leaving few singletons.
Two features stand out.
First, the distributions are long-tailed. Mean family size exceeds the median in every configuration, and the largest families reach 137 variants, so a minority of parents carries a disproportionate share of the mutational variants.
Second, the hyperparameters redistribute units between families rather than into or out of them. Tightening the alignment cut-off from 0.7 to 0.8 more than doubles the number of families while more than halving their mean size. Total family membership, therefore, barely moves, which is why coverage stays above 94\% across configurations.
The alignment cut-off thus strongly controls the family size.
Note that this control is a trade-off between the relatedness of family members and family size.
The same picture holds at 12,800 (Figure~\ref{sup:fig:prelim_kde_a}). Further analyses of substitution matrices and alignment cut-offs are in Section~\ref{sup:sect:hyper_impact}.



Average length of a unit in the vocabulary grows with vocabulary size, from $\sim$2.5 residues at 800 to $\sim$4 at 51,200.
PUMA average unit length tracks BPE closely up to 12,800; only at the largest sizes does PUMA's stay slightly shorter, since additional budget of PUMA goes to variants of existing parents rather than to new, longer merges (Figure~\ref{sup:fig:unitlengths_shared}). 

The vocabulary overlap between PUMA and BPE at 51,200 is only 54\%.
In practice, when we segment a list of protein sequences using these two algorithms, $\sim$90\% of the units that occur exist in both vocabularies (Figure~\ref{sup:fig:shared_units}).
This is what Zipf's law predicts: a small shared set carrying most of the usage, as established for BPE protein units~\citep{suyunu2024linguistic}. PUMA's distinction is therefore not the replacement of those high-frequency units, but rather the replacement of the rarely used long BPE units with a structure of mutational variants.

\subsection{Validating the Biological Significance of PUMA Units}
\label{sec:bio_significance}

We assessed whether the relationships encoded in PUMA families reflect conserved patterns of evolution and function. In these experiments, we evaluated multiple hyperparameter configurations to ensure robustness; for figures containing violin plots, each distribution spans the 55 PUMA configurations trained at that vocabulary size (as detailed in Section~\ref{sup:sect:hyperparameter}).

\subsubsection{PUMA Families Favor Benign Mutations}
\label{sec:benign}
A PUMA family is analogous to a morpheme that occurs in different surface forms across words or sentences. Therefore, we expect that mutations that remain within a PUMA family conserve the behavior (or function) of a protein to a higher degree. We tested this on the expert-annotated ClinVar database~\citep{landrum2018clinvar}, using the ProteinGym~\citep{notin2023proteingym} benchmark, which labels human variants as benign ($\sim$30K) or pathogenic ($\sim$32K). For each single point mutation, we record two events: $P$, the PUMA unit before and after the mutation belongs to the same PUMA family; and $S$, the substitution score is non-negative in the corresponding matrix. 

For the substitutions in this dataset, we count $P$ and $S$ events and for each event type, ask: ``Of all events of that type, what is the ratio of events where this mutation has a benign clinical outcome (i.e.\ benign rate)?''. Due to the slight imbalance between the two labels (30K vs. 32K), benign event counts are artificially upscaled (by $\sim\frac{32}{30}$) to achieve balanced ratios. PUMA proposes mutations only where the substitution scores are already non-negative, so $S$ is the broader condition and itself a decent predictor of variant effect. Enrichment of $P$ over $S$ is therefore directly attributable to PUMA's refinement rather than to the matrix underneath it.    

\begin{figure}[!htbp]
    \begin{center}
    \includegraphics[width=.55\textwidth]{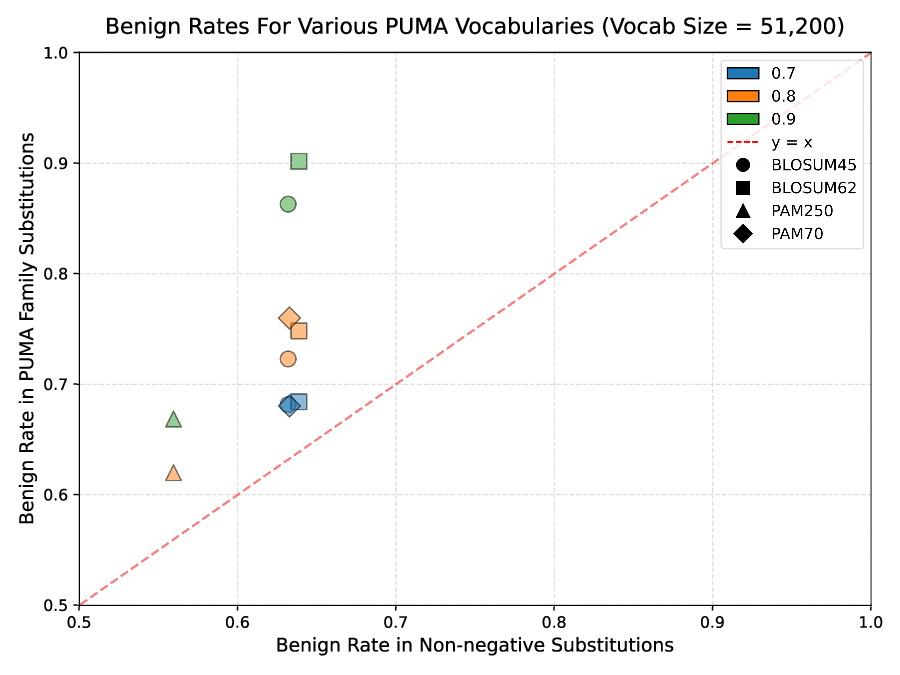}
    \caption{
    Benign rate of PUMA family substitutions ($P$ events) against that of non-negative-scoring substitutions ($S$ events) under the same matrix, at vocabulary size 51,200.
    Colors: alignment cut-off; shapes: substitution matrix; frequency cut-off fixed at 0.05.
    }
    \label{fig:clinvar}
    \end{center}
\end{figure}

In Figure~\ref{fig:clinvar}, benign rates of $P$ and $S$ events are shown across hyperparameter configurations. All throughout, $P$ events carry higher benign rates than $S$ events — by 10\% to 50\% in relative terms — and every point in Figure~\ref{fig:clinvar} is significant against the null of no enrichment ($\mathrm{RR}=1$; two-sided Wald test on $\log \mathrm{RR}$, Benjamini–Hochberg-adjusted $p < 10^{-4}$). Note that $P$ is the rarer event, since sharing a family is a stronger condition than a non-negative score, and the claim here is not about predictive accuracy across the dataset. These results suggest that sharing a PUMA family is a valid biological signal indicating a degree of functional coherence among its members.
%
%
\subsubsection{PUMA Genealogy Improves Functional Representation via Topic Modeling}
\label{sec:topic_modeling}

Members of a PUMA family are linked by plausible amino acid substitutions, so we expect them to contribute similarly to protein function.
A representation of protein function should therefore improve when PUMA family members are treated as related rather than unrelated units.
We test this with a class-based topic representation~\citep{grootendorst2022bertopic}, in which Molecular Function (MF) Gene Ontology (GO) terms are topics and units are words.


From the human UniRef50 dataset and the experimentally validated generic GO-slim of QuickGO~\citep{binns2009quickgo}, we obtained 12 MF terms with roughly the same number of annotated proteins (Section~\ref{sec:tm_method}).
Each GO term $c$ is represented by a vector with one coordinate per unit. All proteins annotated with $c$ are segmented, and $x_{c,t}$ corresponds to the count of occurrences of unit $t$ in these proteins. Class-based TF-IDF (c-TF-IDF) turns these counts into weights $w_{c,t}$ that favor units frequent in $c$ and penalize units common to all GO terms (Section~\ref{sec:tm_method}); the resulting weight vector $\mathbf{w}_c$ is the standard c-TF-IDF representation of GO term $c$.

Standard c-TF-IDF treats units as independent coordinates and ignores the relationships discovered by PUMA. For graph-aware c-TF-IDF, we encode the genealogy as a symmetric weighted adjacency matrix $A$ with one row and one column per unit. Each entry $A_{st}$ is defined as follows:
\begin{equation}
A_{st} =
\begin{cases}
\alpha, & \text{if } s,t \text{ are hierarchical parent and child,}\\
\beta,  & \text{if } s,t \text{ are mutational siblings,}\\
\theta, & \text{if } s,t \text{ are mutational parent and child,}\\
0,      & \text{otherwise.}
\end{cases}
\label{eq:adjacency}
\end{equation}
Before c-TF-IDF is computed, the unit counts of each GO term $c$ are smoothed over this graph,
\begin{equation}
\tilde{x}_{c,t} = (1-\lambda)\,x_{c,t}
+ \lambda\,\frac{\sum_{s} A_{st}\,x_{c,s}}{\sum_{s} A_{st}},
\label{eq:graph_smoothing}
\end{equation}
so each unit keeps a fraction $1-\lambda$ of its own count and receives a fraction $\lambda$ of the weighted mean count of its relatives. 
The graph-aware representation of term $c$, $\tilde{\mathbf{w}}_c$, is obtained by computing the c-TF-IDF weights from $\tilde{x}_{c,t}$ instead of $x_{c,t}$.
BPE records only the hierarchical parent--child link, so graph-aware BPE smooths over the merge tree alone ($\beta=\theta=0$).
Smoothing strength ($\lambda$) and relation weights ($\alpha$, $\beta$, $\theta$) were selected for each tokenizer on a separate set of annotated proteins from those used for reporting (Section~\ref{sec:tm_selection}).

We compared these GO term representations with representations derived from ESM-2 650M~\citep{lin2023evolutionary} embeddings, whose mean-pooled embeddings are widely used as sequence-based functional representations, including for GO-based similarity and function prediction~\citep{ccevrim2025benchmarking}. 
To obtain ESM-2-based GO term representations, each protein is embedded by averaging the final hidden layer of ESM-2, and each GO term is represented by averaging the embeddings of its annotated proteins. For each representation type, we computed the pairwise cosine similarity for all pairs of the 12 GO terms, yielding a vector of 66 similarity values. We then report Spearman's rank correlation between the two resulting 66-dimensional vectors, one from the ESM-2-based representation and one from a c-TF-IDF-based representation.

\begin{figure}[!htbp]
    \begin{center}
    \includegraphics[width=.68\textwidth]{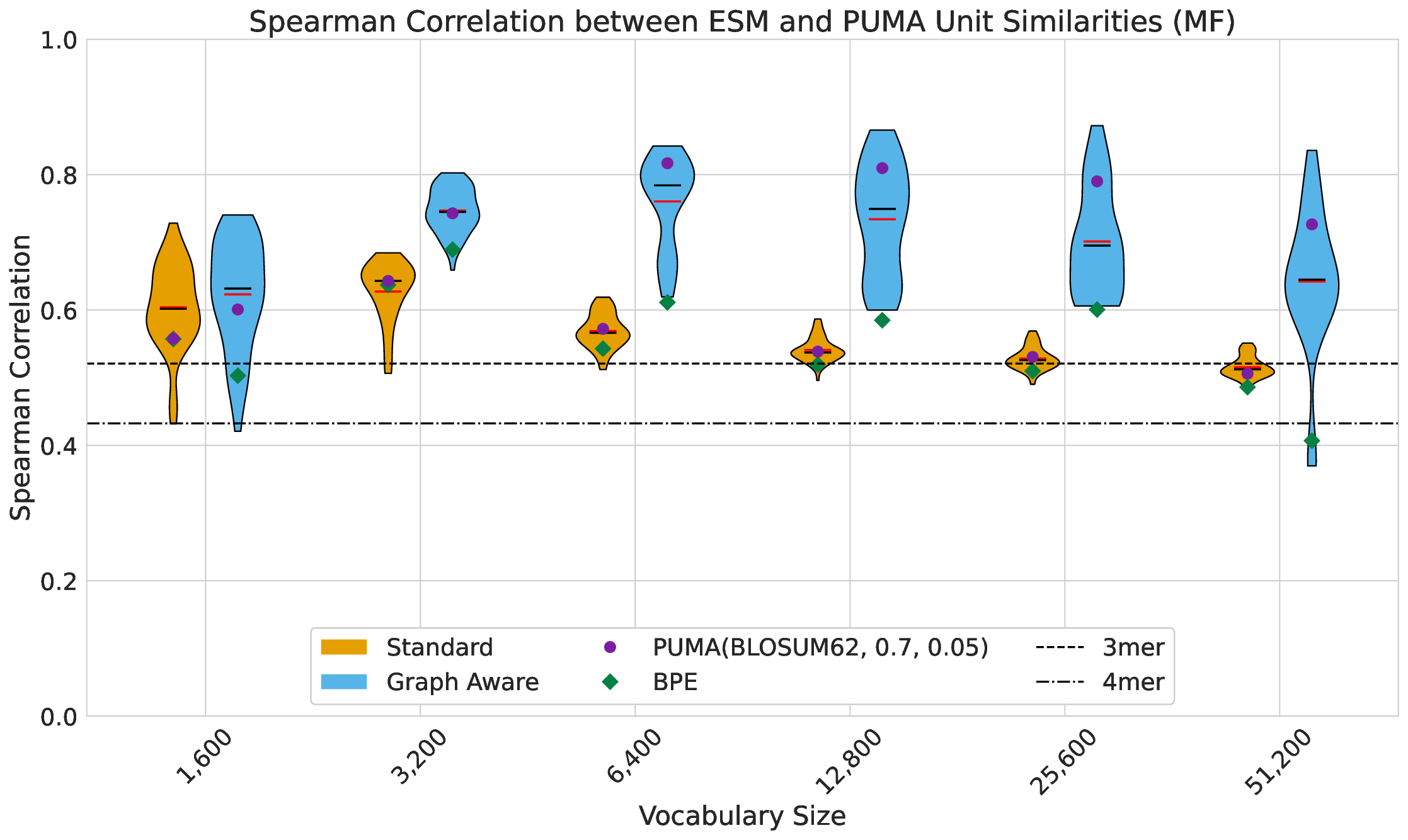}
    \caption{
    Spearman correlation between c-TF-IDF-derived GO vectors and ESM-2-based vectors for Molecular Function, for PUMA, BPE (Standard, Graph-Aware) and k-mer baselines. Violins span all PUMA configurations at a given vocabulary size, each averaged over ten held-out GO-term draws; black and red bars are the median and mean. Vocabularies of 800 are omitted because there the correlation is indistinguishable from zero for every tokenizer ($p \approx 1$).
    }
    \label{fig:go_topic_all}
    \end{center}
\end{figure}

Figure~\ref{fig:go_topic_all} shows the outcome for vocabulary sizes from 1,600 to 51,200; each violin spans the PUMA configurations at that size. Two trends carry the result. Without the genealogy, as the vocabulary grows, the term representation vectors become sparser, standard c-TF-IDF degrades for PUMA and BPE alike, and both approach the flat k-mer reference lines. With the genealogy, graph-aware PUMA is the strongest representation at every vocabulary size and holds its level as the vocabulary grows, so the gap between the two PUMA variants (standard and graph-aware) widens exactly where units are rarest. Graph-aware BPE does not reproduce this: its gain over standard BPE is small and inconsistent, so smoothing over a merge tree is not a substitute for smoothing over a mutational genealogy. A second, independent check points the same way.
Relaxing the alignment cut-off $a$ from 0.9 to 0.7 enlarges the families; it leaves standard c-TF-IDF unchanged yet consistently improves graph-aware PUMA from 6,400 upward (Figure~\ref{sup:fig:go_thrs}), so the gain follows the genealogy.

These results suggest that PUMA families capture information beyond already informative baselines such as the substitution matrix (Section~\ref{sec:benign}) and standard c-TF-IDF.
%
%
%

\subsubsection{Case Study}


As an illustrative example, we conducted a qualitative case study of the \textit{SRGFGFV} family, which comprises relatively long units and was strongly associated with RNA binding in our topic modeling experiments. We examined whether its units capture biologically meaningful local relationships across proteins with diverse functions and evolutionary relatedness.
The family consists of seven units (\textit{SRGFGFV}, \textit{SKGFGFV}, \textit{KRGFGFV}, \textit{SRGFAFV}, \textit{SKGYGFV}, \textit{SKGYGFI}, and \textit{SHGYGFL}) found in RNA- and DNA-binding proteins. 

Our study with \textit{SRGFGFV} family revealed pattern conservation rather than membership in the same protein family. At the closest level of homology, PUMA recovered the same units across paralogs such as hnRNP-A family proteins including HNRNPA1/A2B1/A3/A1L2/A1L3. HNRNPA1 (P09651) and HNRNPA2B1 (P22626) share a conserved structure consisting of two RNA Recognition Motifs (RRMs)~\citep{muto2012structural} with \textit{SRGFGFV} unit followed by a glycine-rich C-terminal region. A second example is the Musashi family proteins MSI1 (O43347) and MSI2 (Q96DH6), which have two RRM regions that contain the \textit{SRGFGFV} unit. 

Importantly, PUMA did not restrict the \textit{SRGFGFV} family to proteins that are only full-length homologs. CIRBP, DAZAP1, PABPC1L, RBFOX, Musashi, and hnRNP-A have distinct overall protein organization and biological roles, and their PUMA units are found within homologous RRM domains. 

On the other hand, \textit{SHGYGFL} unit in SAMD11 proteins A0A087WXB3 and A0A087X223, which contain a sterile alpha motif (SAM) domain but no annotated RRM, is a contrasting case.
SAMD11 is not homologous to any of the RRM-containing proteins with \textit{SRGFGFV} family units, illustrating that local PUMA unit relatedness can be distinguished from protein or domain homology.
%
%
\subsection{PUMA Units as Protein Language-Model Tokenizers}
\label{sec:plm}


A trained PUMA vocabulary enables protein sequences to be segmented into units, allowing it to serve directly as a tokenizer for protein language models (PLMs). In this section, we evaluate the effectiveness of PUMA in this use case.
Our references are BPE, an established tokenizer for protein language models~\citep{tan2024peta,dotan2024effect}, and amino acid (AA) which is the default tokenizer of PLMs.

\subsubsection{Experiment Design}
%
%
Here, we use the complete UniRef50 release (2024-06) instead of the human subset, since pretraining PLMs requires more data and the downstream benchmark is not human-specific. Following the ESM-2 protocol, we partition and decontaminate the dataset before training the tokenizers (Section~\ref{sup:sect:dataset}).

We pretrained masked language models ($\sim$35\,M parameters) from random initialization for one epoch, using the ESM-2 architecture as a fixed reference; no released weights are used. Across runs we vary only the tokenizer and its embedding table. We compare four tokenizers: single-residue amino acids (AA), BPE, PUMA, and its parent-collapsed variant, PUMA\_PC. Each multi-residue tokenizer is trained for seven vocabulary sizes, ranging from 800 to 51{,}200. 
PUMA runs are replicated in three configurations that differ in how much of the vocabulary budget is allocated to mutational variants — BLOSUM62 at alignment cut-offs of 0.7 and 0.8, and PAM70 at 0.7, with a frequency cut-off of 0.05 throughout. The results below are for PUMA (BLOSUM62, 0.7, 0.05), the two other configurations give the same conclusions (Table~\ref{sup:tab:robust}).

PUMA\_PC uses the merge rules of a trained PUMA tokenizer but maps each unit to the mutational parent of its family. Singletons are left unchanged. The segmentation is therefore identical to PUMA, while the embedding table shrinks by 2.1 times at 800 and 12.9 times at 51{,}200 (Table~\ref{sup:tab:conv}).

All models are evaluated on 13 tasks from the PETA benchmark~\citep{tan2024peta}, 18 task/splits in total (Table~\ref{sup:tab:tasks}).
We evaluate in two arms. In PLM-ft, the pretrained encoder is fully fine-tuned on each task. In Scratch, a single-layer encoder is trained on each task without pretraining, so this arm shows what the tokenizer alone contributes. 
Every run uses at least three seeds; protocol and hyperparameters are in Section~\ref{sup:sect:plm}.

\begin{table*}[!t]\centering
\renewcommand{\arraystretch}{0.95}   
\caption{\textbf{Downstream scores on the thirteen PETA tasks.} One split per task (GB1: two-vs-rest; Meltome: mixed; remote homology: fold holdout); all splits are in Tables~\ref{sup:tab:ft} and~\ref{sup:tab:sc}. Spearman's $\rho$ for the variant-effect tasks and Meltome, accuracy otherwise; higher is better. Means over seeds. AA: single-residue baseline (33 tokens). PUMA and PUMA\_PC (PC): BLOSUM62, 0.7, 0.05. PLM-ft: pretrained encoder, fully fine-tuned. Scratch: single-layer encoder trained on each task without pretraining.}
\label{tab:main}\scriptsize\setlength{\tabcolsep}{2pt}
\begin{tabular*}{\textwidth}{@{}l@{\extracolsep{\fill}} *{10}{c}@{\hspace{6pt}}*{10}{c}@{}}
\toprule
& \multicolumn{10}{c}{\textbf{PLM-ft}} & \multicolumn{10}{c}{\textbf{Scratch}}\\
\cmidrule(lr){2-11}\cmidrule(lr){12-21}
 & AA & \multicolumn{3}{c}{800} & \multicolumn{3}{c}{12,800} & \multicolumn{3}{c}{51,200} & AA & \multicolumn{3}{c}{800} & \multicolumn{3}{c}{12,800} & \multicolumn{3}{c}{51,200}\\
\cmidrule(lr){3-5}\cmidrule(lr){6-8}\cmidrule(lr){9-11}\cmidrule(lr){13-15}\cmidrule(lr){16-18}\cmidrule(lr){19-21}
Task & 33 & BPE & PUMA & PC & BPE & PUMA & PC & BPE & PUMA & PC & 33 & BPE & PUMA & PC & BPE & PUMA & PC & BPE & PUMA & PC\\
\midrule
\addlinespace[1.5pt]\multicolumn{21}{@{}l}{\textit{Variant effects}}\\[0.5pt]
Fluorescence & .68 & .68 & .68 & .68 & .68 & .68 & .66 & .68 & .68 & .66 & .30 & .57 & .56 & .52 & .60 & .63 & .61 & .65 & .63 & .62\\
AAV & .81 & .81 & .83 & .81 & .77 & .78 & .73 & .64 & .78 & .71 & .33 & .34 & .09 & .23 & .45 & .58 & .49 & .42 & .52 & .58\\
GB1 & .59 & .39 & .40 & .47 & .50 & .44 & .48 & .52 & .50 & .42 & .29 & .38 & .28 & .37 & .16 & .15 & .05 & .40 & .44 & .37\\
Stability & .77 & .74 & .74 & .75 & .69 & .75 & .69 & .74 & .72 & .71 & .64 & .52 & .54 & .55 & .36 & .50 & .37 & .50 & .46 & .47\\
SolMut CS & .49 & .51 & .51 & .48 & .47 & .50 & .48 & .45 & .50 & .46 & .04 & .15 & .08 & .13 & .14 & .22 & .15 & .23 & .26 & .22\\
SolMut BLAT & .52 & .54 & .50 & .51 & .46 & .49 & .47 & .48 & .48 & .46 & .09 & .12 & .11 & .09 & .17 & .12 & .17 & .17 & .23 & .17\\
SolMut LGK & .58 & .55 & .55 & .54 & .51 & .52 & .54 & .51 & .52 & .52 & \makebox[0pt][r]{$-$}.01 & .06 & .12 & .11 & .16 & .12 & .09 & .08 & .09 & .12\\
\addlinespace[1.5pt]\multicolumn{21}{@{}l}{\textit{Protein properties}}\\[0.5pt]
Meltome & .61 & .60 & .61 & .59 & .57 & .60 & .54 & .58 & .59 & .54 & .50 & .46 & .45 & .45 & .39 & .39 & .39 & .35 & .35 & .36\\
DeepSol & .73 & .73 & .72 & .74 & .72 & .73 & .74 & .73 & .72 & .72 & .51 & .64 & .57 & .57 & .57 & .64 & .60 & .63 & .63 & .64\\
DeepLoc-2 & .93 & .92 & .92 & .92 & .92 & .92 & .92 & .92 & .92 & .92 & .90 & .89 & .89 & .90 & .88 & .88 & .89 & .88 & .88 & .89\\
DeepLoc bin. & .91 & .91 & .91 & .91 & .90 & .91 & .89 & .91 & .91 & .90 & .75 & .74 & .72 & .73 & .72 & .71 & .74 & .70 & .67 & .72\\
\addlinespace[1.5pt]\multicolumn{21}{@{}l}{\textit{Homology and interaction}}\\[0.5pt]
Remote homology & .25 & .22 & .22 & .20 & .19 & .20 & .18 & .20 & .19 & .17 & .07 & .09 & .10 & .09 & .09 & .09 & .07 & .08 & .09 & .10\\
PPI SHS27k & .53 & .53 & .53 & .53 & .54 & .53 & .53 & .53 & .54 & .53 & .55 & .56 & .56 & .55 & .57 & .56 & .55 & .57 & .56 & .56\\
\bottomrule
\end{tabular*}
\end{table*}

\subsubsection{Experiment Results}

Table~\ref{tab:main} reports the results of the tokenizers for one split of each of the $13$ tasks. The results for all splits and vocabulary sizes follow the same trend and are provided in Tables~\ref{sup:tab:ft} and~\ref{sup:tab:sc}. In both arms (PLM-ft and Scratch), PUMA and BPE perform comparably across tasks and vocabulary sizes.
Differences on individual tasks occur in both directions, and a paired Wilcoxon signed-rank test over the 13 tasks finds no significant difference between the two tokenizers, except at vocabulary size 12{,}800 under PLM-ft, where PUMA holds a significant advantage (Section~\ref{sup:sect:paired}).
The main finding of this section is that, although PUMA allocates most of its vocabulary to rare mutational variants that BPE would not retain, its downstream performance is comparable to BPE.

In PLM-ft, all multi-residue tokenizers trail AA. This is a known property of multi-residue tokenizers on pretrained encoders~\citep{tan2024peta} rather than a property of PUMA: BPE, PUMA and PUMA\_PC lose by similar margins and on the same tasks. Most of the gap comes from remote homology and some of the variant-effect tasks, while the protein-property tasks stay close to AA.


In scratch (i.e., without pretraining), all multi-residue tokenizers outperform AA on most variant-effect tasks, often by wide margins, and remain competitive on most protein-property tasks. The advantage is not explained by an increase in parameter count: PUMA\_PC has a much smaller embedding table than PUMA at the same segmentation and performs similarly. One possible reason is that a multi-residue vocabulary itself carries local sequence context that the AA encoder would otherwise have to learn from the task data alone. Consistent with this, the advantage disappears once the encoder gains the ability to better represent local context via pretraining.

PUMA\_PC trades a small loss in accuracy for efficiency. It scores slightly below PUMA in PLM-ft, most visibly at the largest vocabularies, and matches it in Scratch, while its embedding table is up to 12.9 times smaller. The smaller table matters mainly for pretraining. Every multi-residue tokenizer pretrains faster than AA because it produces shorter token sequences. However, the masked-language-model head projects onto the full vocabulary at every masked position, and this cost grows with the vocabulary. For BPE and PUMA, the speedup over AA therefore peaks at about 1.7 times at intermediate sizes and falls to 1.1 times at 51{,}200. PUMA\_PC keeps PUMA's segmentation but projects onto the collapsed vocabulary, so its speedup keeps growing, to 1.8 times at 51{,}200, where it pretrains 1.7 times faster than PUMA (Table~\ref{sup:tab:pretrain_time}). Fine-tuning has no vocabulary-sized head, so every multi-residue tokenizer fine-tunes 1.4--1.8 times faster than AA, and the speedup rises with vocabulary size (Table~\ref{sup:tab:ft_time}).

\section{Conclusion and Discussion}


We presented PUMA, an algorithm that learns a vocabulary of protein units directly from sequence. Motivated by the fact that proteins evolve through mutation, PUMA employs a mutation-aware merging procedure that uses a substitution matrix to identify mutational variants of each newly discovered unit.
Each unit is thus linked to its variants, resulting in both a vocabulary of protein units and a genealogy of their relationships.

Our validations demonstrate that this genealogy reflects genuine biological organization. Mutations that keep a unit within its PUMA family are enriched for benign clinical outcomes beyond what the underlying substitution matrix already predicts. 
Incorporating the PUMA genealogy into graph-aware topic modeling improves molecular function representations.
An equivalently tuned graph-aware BPE shows no comparable gain, indicating that the benefit is specific to the genealogy.
%
%
Tracing the \textit{SRGFGFV} family illustrates what PUMA families look like at the level of individual proteins. Units from this family recur across RNA-binding proteins that are not close homologs, and one occurs in SAMD11 proteins, which lack an annotated RRM domain. 
This suggests that PUMA identifies local sequence-level relatedness beyond conventional protein or domain homology.
%
Masked language models pretrained on PUMA-tokenized sequences perform on par with BPE across thirteen downstream tasks, 
despite spending most of their vocabulary budget on mutational variants a frequency-driven tokenizer would not retain. Collapsing each family onto its mutational parent converts the family structure into a compression mechanism, shrinking the embedding table by up to 12.9 times and accelerating pretraining.
%
Neither multi-residue tokenizer outperforms the single-residue baseline when applied to a pretrained encoder, consistent with previous reports; nevertheless, performance is comparable across approaches. The PUMA genealogy further provides a natural basis for developing post-hoc explainability methods for protein language models trained with PUMA tokenizers.

While our biological analyses were confined to human proteins by the availability of clinical and functional ground truth, applying PUMA across diverse taxa could establish a field of comparative proteo-linguistics, revealing which protein units are universal building blocks and which have evolved as lineage-specific dialects.
Another promising direction is to integrate PUMA genealogy directly into neural architectures. Attention mechanisms or graph neural networks could be biased to leverage these related sibling units, bridging ground-up representation with deep learning’s predictive power.

PUMA offers a complementary perspective on protein analysis. While large-scale models prioritize predictive accuracy and post-hoc methods aim to interpret them, PUMA builds an explanatory foundation from first principles.
PUMA's biologically grounded units, together with their genealogy, may enable a deeper understanding of the language of life.

\section*{Author Contributions}

B.S., Ö.D., and A.Ö. conceived and designed the study. B.S. and Ö.D. implemented the algorithms, conducted the experiments, and wrote the manuscript.
B.S., Ö.D., and A.Ö. analyzed the results. I.A.O. and H.K.B. evaluated the PUMA units and families for biological relevance. A.Ö. supervised the project. All authors reviewed the manuscript.

\section*{Supplementary data}
The Supplementary Material is appended to this preprint (Sections~\ref{sup:sect:dataset}--\ref{sup:sect:plm}).

\section*{Conflict of interests}
No competing interest is declared.

\section*{Funding}
This work is supported by ERC grant (LifeLU, 101089287). Views and opinions expressed are however those of the author(s) only and do not necessarily reflect those of the European Union or the European Research Council Executive Agency. Neither the European Union nor the granting authority can be held responsible for them.

\section*{Data and Code Availability}
All data are from the public sources UniProt, ProteinGym, QuickGO, and the PETA benchmark.
Pretrained ESM-2 weights are available on Hugging Face.
Source code and trained PUMA models are available at \mbox{\url{https://github.com/boun-tabi-lifelu/PUMA}}.


\bibliographystyle{oup-abbrvnat}
\bibliography{reference_v3}

\clearpage
\makeatletter
\setcounter{section}{0}
\setcounter{table}{0}
\setcounter{figure}{0}
\setcounter{algorithm}{0}
\setcounter{equation}{0}
\renewcommand\thesection{S\@arabic\c@section}
\renewcommand\thetable{S\@arabic\c@table}
\renewcommand\thefigure{S\@arabic\c@figure}
\renewcommand\thealgorithm{S\@arabic\c@algorithm}
\renewcommand\theequation{S\@arabic\c@equation}
\def\theHsection{S\@arabic\c@section}
\def\theHtable{S\@arabic\c@table}
\def\theHfigure{S\@arabic\c@figure}
\def\theHalgorithm{S\@arabic\c@algorithm}
\def\theHequation{S\@arabic\c@equation}
\makeatother
\renewcommand{\arraystretch}{1.06}

\begin{center}
{\Large\bfseries Supplementary Material}
\end{center}
\vspace{-10pt}

\section{Dataset}
\label{sup:sect:dataset}

The vocabulary analyses and biological validations use the human UniRef50 subset described in Section~\ref{sec:experiments}. The language-model experiments instead use the complete UniRef50 release (2024-06), partitioned and decontaminated following the protocol used for ESM-2~\citep{lin2023evolutionary}.

The corpus was partitioned by randomly selecting 0.5\% ($\approx$250{,}000) of sequences to form the validation set. Training sequences were then filtered following the procedure of \citet{meier2021language}: an MMseqs2~\citep{steinegger2017mmseqs2} search is run with the training set as query database and the validation set as target database, using

\begin{center}
\texttt{-{}-min-seq-id 0.5 -{}-alignment-mode 3 -{}-max-seqs 300 -s 7 -c 0.8 -{}-cov-mode 0}
\end{center}

\noindent and all training sequences matching a validation sequence at 50\% sequence identity under this search are removed from the training set. Because the same corpus is shared by every tokenizer, this decontamination keeps the comparison between AA, BPE, PUMA and PUMA\_PC free of tokenizer-specific leakage.

Fitting a tokenizer does not require the full corpus. Merge statistics converge well before the entire training split is consumed, and PUMA's cost is dominated by the number of distinct pairs whose mutational neighborhoods must be searched, which grows with corpus size while the resulting vocabulary does not change materially. We therefore build all tokenizers on a random sample of 2M sequences drawn from the decontaminated training split. BPE and PUMA are fitted on the same sample, so any difference between the resulting vocabularies is attributable to the algorithm rather than to the data.

\section{PUMA Training Algorithm}
\label{sup:sect:algorithm}
Algorithm~\ref{sup:algo:PUMA} gives the pseudocode for the discovery process described in Section~\ref{sec:puma}.

\begin{algorithm}
\caption{PUMA Training}
\begin{algorithmic}[1]
\Require Dataset $D$ with protein sequences, Vocabulary size $V$, Substitution matrix $S$, Alignment score cut-off $a$, Frequency cut-off $f$
\Ensure Vocabulary $V_{PUMA}$, Ruleset $R_{PUMA}$
\State Initialize $V_{PUMA}$ as set of standard amino acids in $D$
\State Initialize max-heap $H$ with frequencies of all pairs in $D$
\While{$|V_{PUMA}| < V$}
    \State Extract most frequent pair $(x, y)$ from $H$
    \State $r: p \leftarrow Merge(x, y)$
    \State Add $p$ to $V_{PUMA}$ and Add $r$ to $R_{PUMA}$
    \For{each amino acid $a_i$ in $p$}
        \State Find substitutions $a_j$ such that $S(a_i, a_j) \geq 0$
    \EndFor
    \State Generate all substitution sequences $p'$ of $p$ using $a_j$
    \For{each $p'$}
        \State Compute alignment score, $score(p, p')$
        \If{$score(p, p') \geq a$}
            \State Find $r': p' \leftarrow Merge(x', y')$
            \State Compute frequency of $p'$ in $D$, $freq(p')$
            \If{$freq(p') \geq f \times freq(p)$}
                \State Add $p'$ to $V_{PUMA}$ and Add $r'$ to $R_{PUMA}$
                \State Add $p' \leftarrow Mutate(p)$ to $R_{PUMA}$
            \EndIf
        \EndIf
    \EndFor
    \State Update $H$ with new pairs and their frequencies
\EndWhile
\end{algorithmic}
\label{sup:algo:PUMA}
\end{algorithm}

\section{PUMA Hyperparameter Configurations}
\label{sup:sect:hyperparameter}

We explored a hyperparameter grid to analyze how evolutionary constraints influence vocabulary construction. The configuration space included:
\begin{itemize}
    \item \textbf{Vocabulary Size:} Ranging from 800 to 51,200, doubling at each step.
    \item \textbf{Substitution Matrices:} BLOSUM62, BLOSUM45, PAM70, and PAM250.
    \item \textbf{Alignment Score Cut-off ($a$):} Tested at 0.7, 0.8, and 0.9. Values below 0.7 were excluded due to exponential search complexity, while PAM250 was restricted to 0.8 and 0.9 to maintain computational feasibility. As $a$ approaches 1, PUMA converges toward standard BPE behavior.
    \item \textbf{Frequency Cut-off ($f$):} Cut-off values of 0, 0.005, 0.05, 0.1, and 0.2 were applied to filter coincidental patterns.
\end{itemize}

Mutations for units shorter than 3 and longer than 12 are not generated. The former is because almost all of the 2 residue units were already getting included in vocabularies without the mutations, whereas the latter is due to the exponential complexity of mutation search. The specific effects of these parameters and their optimal configurations for biological validity are analyzed in the subsequent sections. A model trained under a particular configuration may be referred to in the format PUMA(Matrix, $a$, $f$).

\section{Impact of PUMA Hyperparameters}
\label{sup:sect:hyper_impact}

\subsection{Impact of Alignment Score Cut-off on Vocabulary}
\label{sup:sect:align_impact}

\begin{figure}[tbp]
    \begin{center}
    \subfloat[]{\includegraphics[width=.4\textwidth]{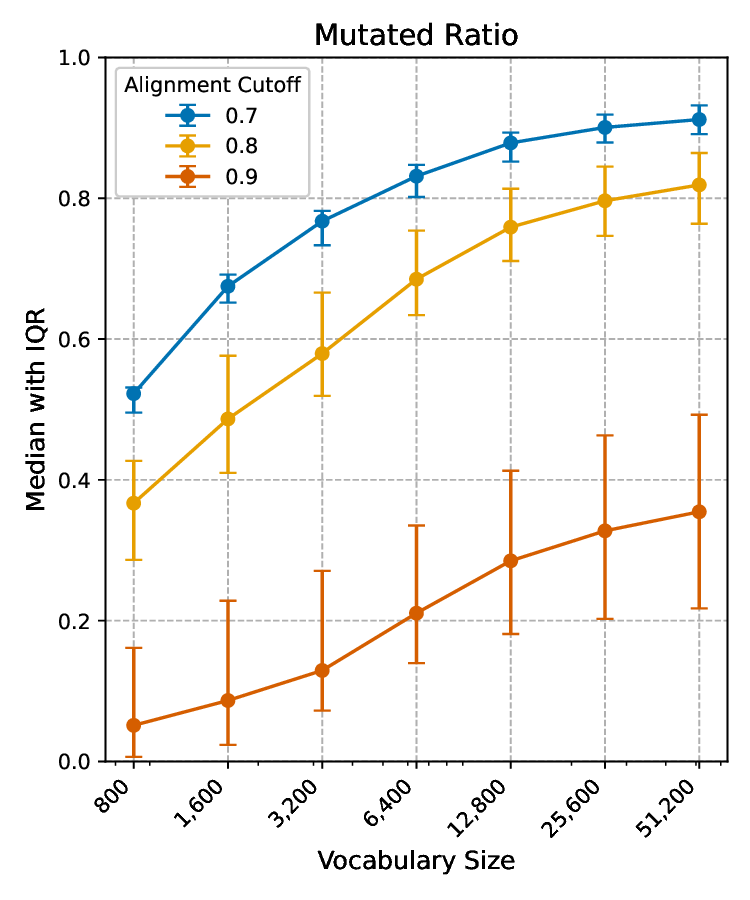}\label{sup:subfig:mutated_ratio_a}}\hspace{.04\textwidth}
    \subfloat[]{\includegraphics[width=.4\textwidth]{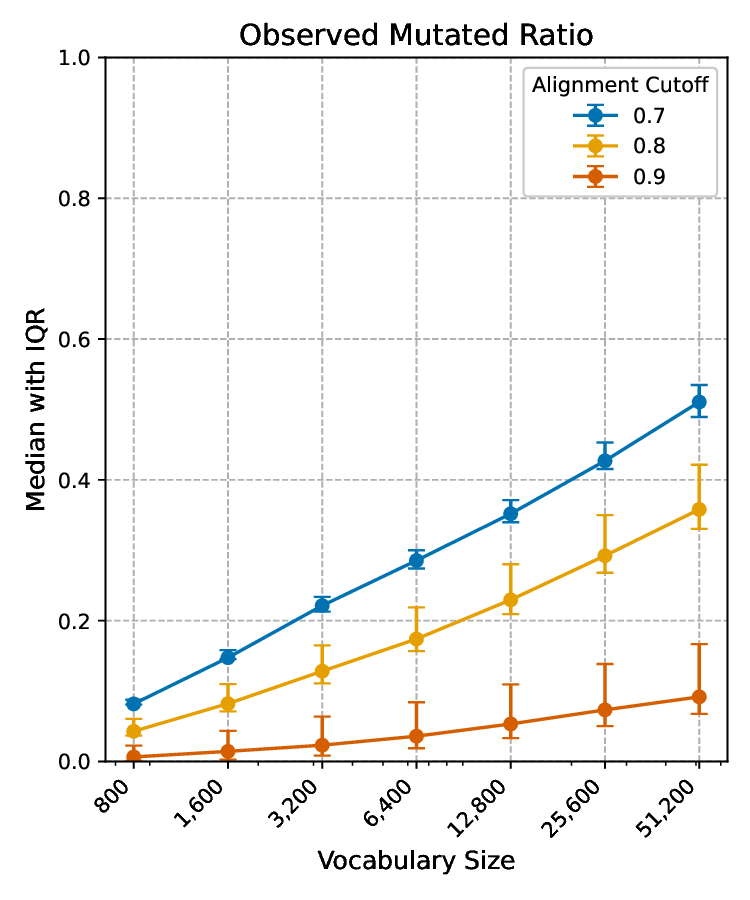}\label{sup:subfig:mutated_ratio_b}}
    \caption{Mutated ratio statistics for PUMA models. Each data point is the interquartile range (IQR) over the PUMA models trained with that alignment cut-off: 20 models (4 substitution matrices $\times$ 5 $f$ values) at $a = 0.8$ and $0.9$, and 15 at $a = 0.7$, where PAM250 is not trained. (a) IQR of the ratio of mutated units to all units in the vocabulary. (b) IQR of the ratio of observed mutated units to all observed units in the dataset (after segmentation).}
    \label{sup:fig:mutated_ratios}
    \end{center}
\end{figure}

The alignment score cut-off ($a$) critically dictates mutational diversity as demonstrated in Figure~\ref{sup:subfig:mutated_ratio_a}. A strict cut-off of 0.9 yields conservative vocabularies resembling standard BPE, whereas relaxing $a$ to 0.8 triggers a substantial expansion in the number of discovered variants. Further relaxing the cut-off to 0.7 yields a more gradual increase in diversity compared to the jump observed at 0.8. For this reason, we focused more on the models with 0.7 and 0.8 $a$ values in our further validations. Notably, while mutations may constitute up to 85\% of the vocabulary, their observation in segmented sequences is approximately 45\% as shown in Figure~\ref{sup:subfig:mutated_ratio_b}. This indicates that while PUMA provides a vast library of mutational variants, the parent protein units still account for the majority of sequence coverage.

\subsection{Impact of Substitution Matrices on Vocabulary}
\label{sup:sect:subs_impact}

We deliberately confine our experiments to the BLOSUM and PAM families. Substitution matrices derived from predicted structures have recently become available, but a systematic comparison of classical and structure-derived matrices reports that the different families behave very similarly in practice~\citep{buschmann2026much}. The identity of the matrix family is therefore unlikely to be the dominant factor in PUMA's behavior; what matters is that some evolutionary prior guides the mutation step. What does remain to be quantified is the effect of the specific matrix chosen within these families, which is the subject of this section.

The choice of substitution matrix (BLOSUM62, BLOSUM45, PAM70, PAM250) affects the content of the discovered units. This has been examined by quantifying the pairwise vocabulary identities (at a vocabulary size of 51200). Vocabulary identity simply corresponds to the proportion of the number of shared units between the two models to the vocabulary size. 

\begin{figure}[tbp] 
    \centering
    \includegraphics[width=.62\textwidth]{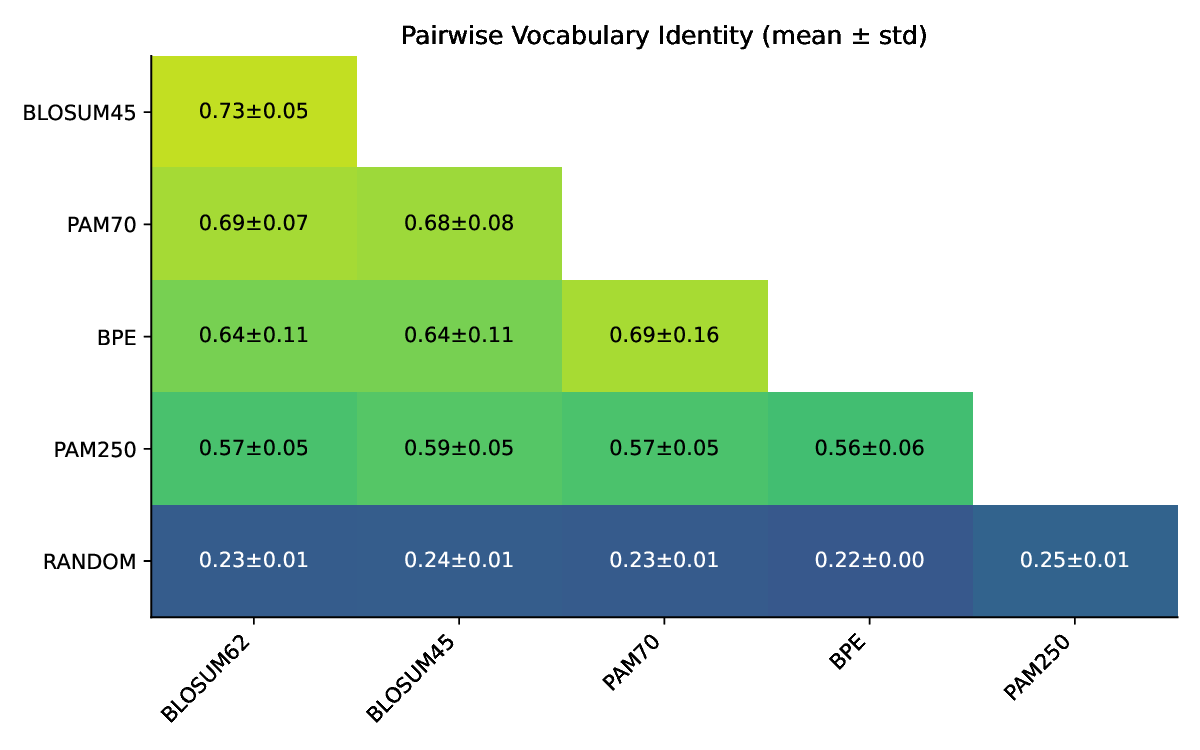}
    \caption{Mean vocabulary identity (proportion of shared units) between RANDOM, BPE, and PUMA for vocabulary size of 51200.}
    \label{sup:fig:vocab_identity}
\end{figure}

Figure~\ref{sup:fig:vocab_identity} reveals that PUMA models trained with different matrices produce notably distinct vocabularies, with mean identity values ranging from 0.56 to 0.73 between models. PAM250 generating the lowest identity value is expected, as it is the matrix that punishes mutations the least. The pairwise identities between the RANDOM baseline and all other models are low ($\sim$23\%) and consistent. In contrast, BPE shares a much higher proportion of its vocabulary with PUMA models ($\sim$64\%). This suggests that both BPE and PUMA capture a shared core of biologically non-random units, while the specific choice of substitution matrix in PUMA allows the model to learn a distinct dialect of mutational variants that extends beyond this shared core.

We performed comparisons across four distinct categories:
\begin{itemize}
    \item \textbf{PUMA vs. PUMA:} We compared vocabularies generated with identical hyperparameters but different substitution matrices.
    \item \textbf{PUMA vs. BPE:} We compared every hyperparameter configuration of PUMA (across all matrices) against the single standard BPE configuration.
    \item \textbf{RANDOM vs. PUMA:} For each PUMA vocabulary, we generated a random vocabulary with an identical unit length distribution by uniformly sampling from standard amino acids.
    \item \textbf{RANDOM vs. BPE:} We compared BPE against 10 independently generated random vocabularies.
\end{itemize}


\begin{figure}[tbp]
    \begin{center}
    \includegraphics[width=.62\textwidth]{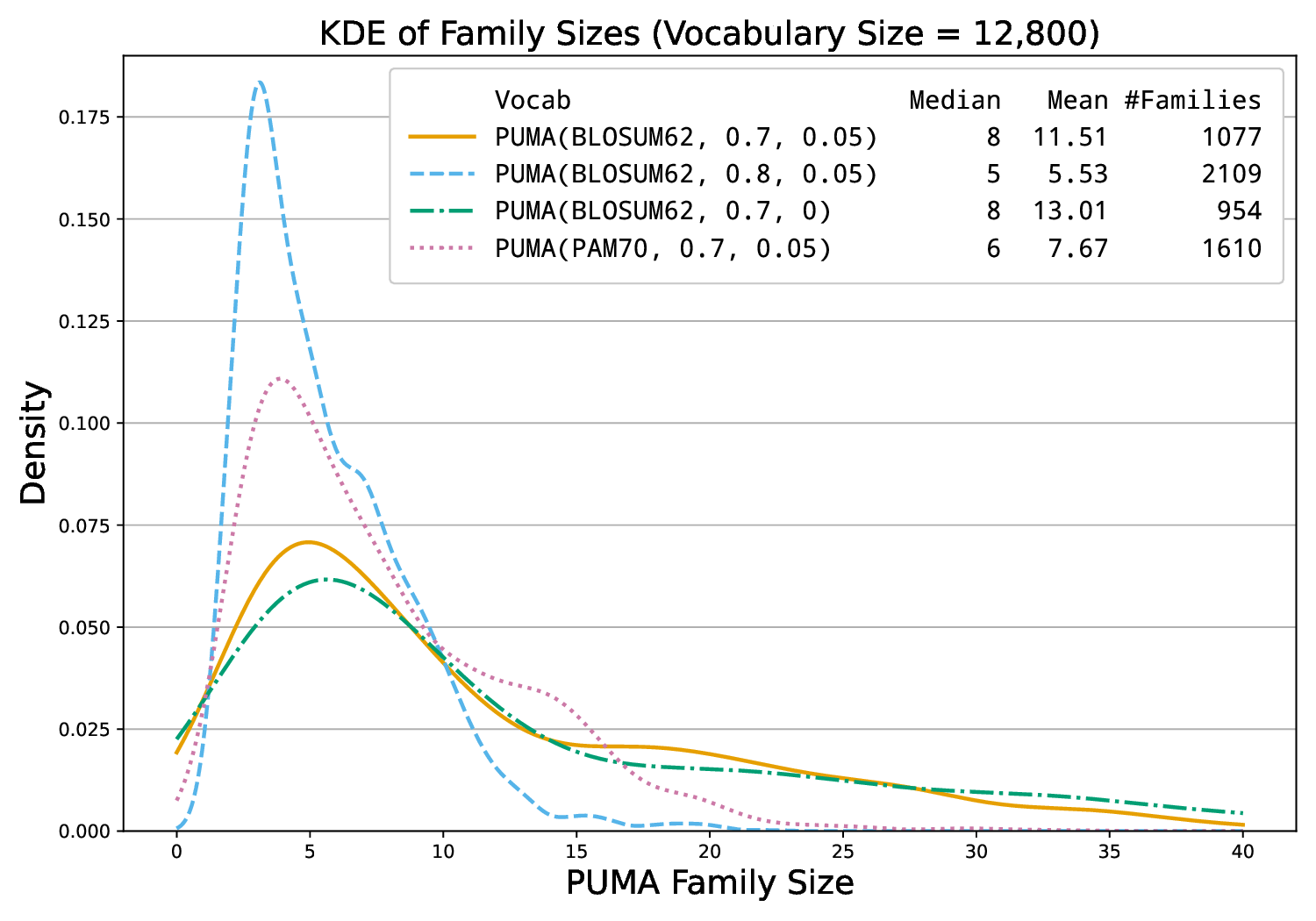}   
    \caption{Distribution of PUMA family sizes at vocabulary size 12,800 for the four configurations of Figure~\ref{fig:prelim_kde}.}
    \label{sup:fig:prelim_kde_a}
    \end{center}
\end{figure}

\begin{figure}[tbp]
    \begin{center}
    \subfloat[]{\includegraphics[width=.4\textwidth]{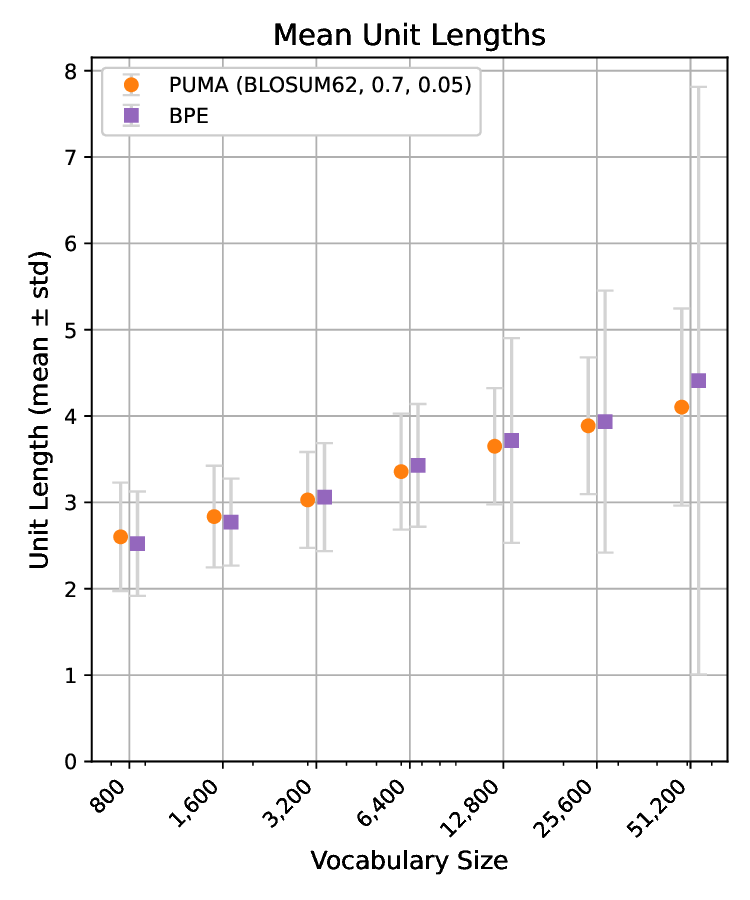}\label{sup:subfig:unitlength_a}}\hspace{.04\textwidth} 
    \subfloat[]{\includegraphics[width=.4\textwidth]{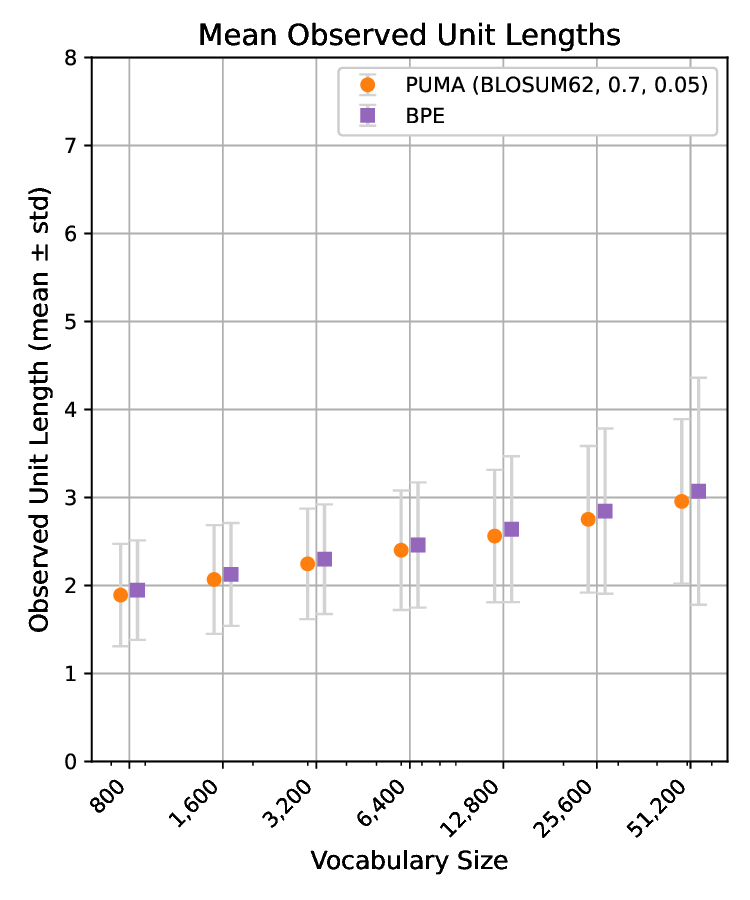}\label{sup:subfig:unitlength_b}}
    \caption{(a)~Mean unit lengths of PUMA and BPE vocabularies at different sizes. (b)~Mean observed unit lengths in PUMA and BPE vocabularies at different sizes. Observed units are units that have been observed after segmenting the dataset.}
    \label{sup:fig:unitlengths_shared}
    \end{center}
\end{figure}

\begin{figure}[tbp]
    \begin{center}
    \includegraphics[width=.62\textwidth]{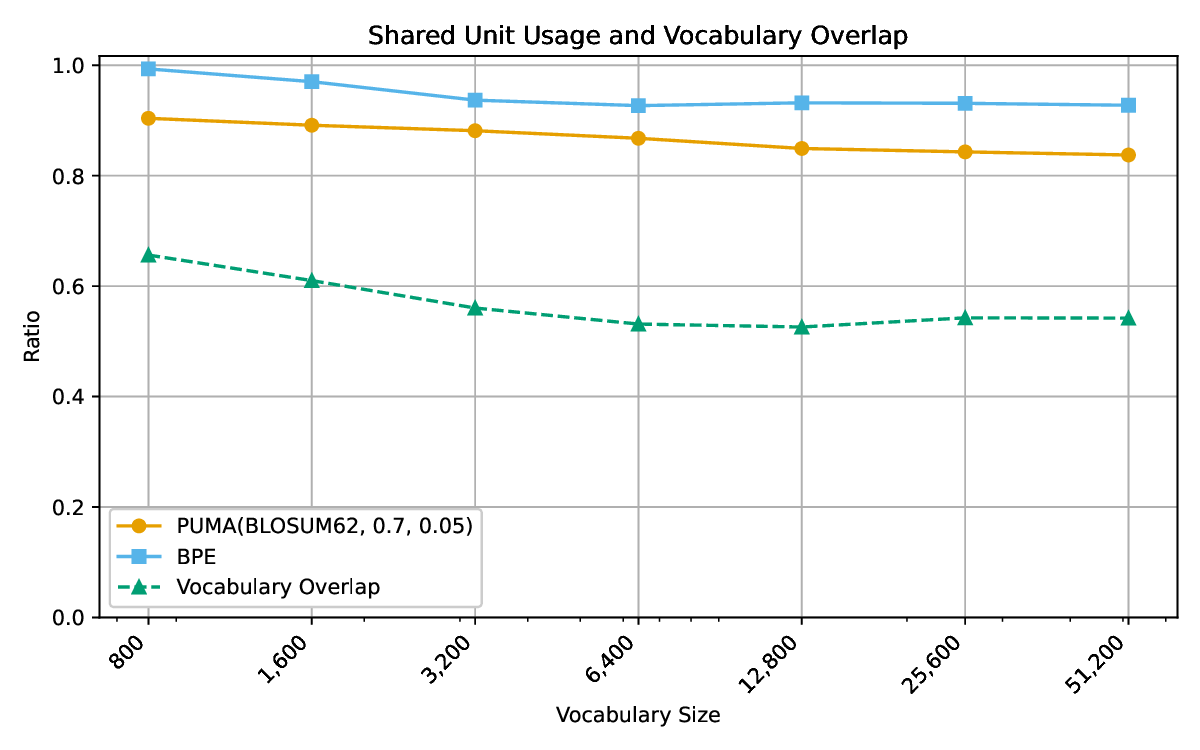}
    \caption{Vocabulary overlap (in green) shows number of units that exist in both vocabularies divided by vocabulary size. Shared unit usage (in yellow and blue) is the ratio of total number of times overlapping units are used divided by total number of units that are used (by PUMA and BPE respectively) in a segmented dataset.}
    \label{sup:fig:shared_units}
    \end{center}
\end{figure}

\section{Graph-aware topic modeling}
\label{sup:sect:topic_modeling}

This section gives the full protocol behind Section~\ref{sec:topic_modeling} and Figure~\ref{fig:go_topic_all}: how the GO-term set, the term vectors and the genealogy graph are built (Section~\ref{sec:tm_method}), how the smoothing hyperparameters were selected and how results are reported (Section~\ref{sec:tm_selection}), and an ablation of the alignment score cut-off that tests whether the gain is carried by family structure (Section~\ref{sec:tm_cutoff}).

\subsection{Method}
\label{sec:tm_method}

\paragraph{GO-term set.}
Proteins are drawn from the human UniRef50 dataset of Section~\ref{sec:experiments}, restricted to sequences shorter than 1,000 residues, and annotated with Molecular Function (MF) terms from the generic GO-slim subset of QuickGO \citep{binns2009quickgo}, keeping only experimental evidence codes (EXP, IDA, IPI, IMP, IGI, IEP).
Term membership in GO is highly imbalanced, so terms with fewer than 100 annotated proteins were excluded and the remaining terms were undersampled by random draw, giving 12 unique MF terms with $222 \pm 38$ annotated proteins per term. We write $P_c$ for the proteins of a draw annotated with term $c$; a protein annotated with several of the selected terms belongs to each corresponding $P_c$. Because the set depends on the draw, every quantity below is computed within a single draw and repeated over multiple independent draws (Section~\ref{sec:tm_selection}).

\paragraph{Term count vectors.}
Following BERTopic~\citep{grootendorst2022bertopic}, the proteins of each term are segmented with the tokenizer and concatenated into one document per term. The count vector $\mathbf{x}_c$ of term $c$ has one entry per unit, and $x_{c,t}$ is the number of occurrences of unit $t$ in the document of $c$, i.e.\ across all segmented proteins of $P_c$. Units occurring in more than 90\% of the term documents, i.e.\ in 11 or 12 of the 12 terms, carry no information about which term they belong to and are discarded (BERTopic's \texttt{max\_df}\,=\,0.9). The coordinates of every term vector are therefore the remaining units of the vocabulary. The same construction is applied to PUMA, BPE and $k$-mer vocabularies; for $k$-mers, the units are non-overlapping windows of length $k$.

\paragraph{c-TF-IDF weighting.}
The weight of unit $t$ in term $c$ is
\begin{equation}
  w_{c,t} = \sqrt{\frac{x_{c,t}}{\sum_{t'} x_{c,t'}}} \cdot \log\!\left(1 + \frac{\bar{N}}{\sum_{c'} x_{c',t}}\right),
  \label{eq:ctfidf}
\end{equation}
where $c'$ runs over the 12 terms and $\bar{N}$ is the mean total unit count per term. The first factor is the relative frequency of $t$ within term $c$, with the square root damping the most frequent units; the second down-weights units that are frequent across all terms. This is BERTopic's \texttt{ClassTfidfTransformer} with \texttt{reduce\_frequent\_words} enabled. The vector $\mathbf{w}_c = (w_{c,t})_t$ is the standard c-TF-IDF representation of term $c$.

\paragraph{Genealogy graph and graph-aware weighting.}
The graph has one node per unit retained above and is encoded by the symmetric adjacency matrix $A$ of Equation~\eqref{eq:adjacency}. A hierarchical edge (weight $\alpha$) links a unit to each of the two units whose merge created it; only direct merges are recorded, not the full merge tree above them. Within a PUMA family, a mutational edge links the mutational parent to each of its children (weight $\theta$), and a sibling edge links every pair of children of the same parent (weight $\beta$). $A$ has no self-loops.

Equation~\eqref{eq:graph_smoothing} can be written in matrix form as
\begin{equation}
  \tilde{\mathbf{x}}_c = \mathbf{x}_c \bigl((1-\lambda) I + \lambda \hat{A}\bigr),
  \qquad
  \hat{A}_{st} = \frac{A_{st}}{\sum_{s'} A_{s't}},
  \label{eq:smoothing}
\end{equation}
where $\mathbf{x}_c$ is a row vector, $I$ is the identity and $\hat{A}$ is $A$ with each column normalized to sum to one. The parameter $\lambda \in [0,1]$ interpolates between the unsmoothed counts ($\lambda = 0$) and pure neighborhood pooling ($\lambda = 1$). The normalization makes the neighborhood term a weighted mean rather than a sum, so the evidence a unit receives from its relatives does not grow with the size of its family, and the relation weights $\alpha$, $\beta$, $\theta$ only set the relative influence of the three relation types. A unit without relatives in the graph (for instance, a singleton whose merge constituents were also discarded) receives nothing from the neighborhood term and keeps $(1-\lambda)\,x_{c,t}$. The graph-aware representation $\tilde{\mathbf{w}}_c$ is obtained by replacing $x$ with $\tilde{x}$ throughout Equation~\eqref{eq:ctfidf}, including $\bar{N}$ and the cross-term sum; everything downstream is unchanged.

BPE records only the merge that created each unit, i.e.\ the hierarchical parent--child relation. Its graph-aware variant therefore uses Equation~\eqref{eq:adjacency} with only $\alpha$ non-zero, and it smooths over the merge tree alone. This is the control for whether the benefit comes from graph-aware smoothing in general or from the mutational relations specific to PUMA. $k$-mer vocabularies have no genealogy and enter only with the standard representation.

\paragraph{Excluded vocabulary size.}
Vocabularies of 800 units are omitted from all comparisons. At that size the Spearman correlation is indistinguishable from zero for every tokenizer ($p \approx 1$), with or without smoothing, so the cell carries no information about the relative merit of the vocabularies.

\subsection{Hyperparameter selection and reporting}
\label{sec:tm_selection}

The smoothing parameters and the resulting correlations both depend on which GO-term draw is used, so selection and reporting were kept on independent draws.

\paragraph{Grid.}
For PUMA, the relation weights and smoothing strength of Equations~\eqref{eq:adjacency}--\eqref{eq:graph_smoothing} range over $\alpha, \beta, \theta \in \{1, 2, 3\}$ and $\lambda \in \{0.1, 0.5, 0.9\}$, giving 81 settings. For graph-aware BPE, which carries only $\alpha$, the grid is $\alpha \in \{1, 2, 3\}$ and $\lambda \in \{0.1, 0.5, 0.9\}$, giving 9 settings.

\paragraph{Selection.}
Ten independent GO-term draws were used for selection only. For each candidate setting we computed its rank among the grid within every tokenizer--vocabulary cell, averaged over the ten draws, and then averaged those ranks across cells. The setting with the lowest mean within-model rank was retained. The rule was applied identically to PUMA and to graph-aware BPE, so neither method is compared against an untuned version of the other. The selected settings are $\lambda = 0.9$, $\alpha = 2$, $\beta = 2$, $\theta = 3$ for PUMA and $\lambda = 0.9$, $\alpha = 3$ for BPE. The selected $\lambda$ is the strongest smoothing in the grid for both tokenizers.

\paragraph{Reporting.}
The selected settings were then frozen, and every correlation in Figure~\ref{fig:go_topic_all} and in Figure~\ref{sup:fig:go_thrs} was computed on ten further GO-term draws, independent from the ten used for selection. No result is reported on a draw that contributed to choosing the parameters under which it was obtained. Each tokenizer--vocabulary cell is summarized by its mean correlation over the ten reporting draws.

\paragraph{What the violins show.}
The PUMA distributions in Figure~\ref{fig:go_topic_all} are taken over PUMA configurations at a given vocabulary size, not over GO-term draws. Each configuration (substitution matrix, alignment score cut-off, frequency cut-off; Section~\ref{sup:sect:hyperparameter}) is one observation, already averaged over the ten reporting draws, so the violins show how sensitive the result is to PUMA's hyperparameters with sampling variation averaged out. BPE and $k$-mers have a single configuration per size and are shown as points or lines.

\subsection{Effect of the alignment score cut-off}
\label{sec:tm_cutoff}

The alignment score cut-off $a$ determines which candidate variants are admitted into a family (Section~\ref{sup:sect:align_impact}) and therefore how much genealogy there is to smooth over. Changing $a$ also changes the vocabulary, but if the graph-aware gain is carried by family structure, it should move graph-aware results while leaving standard c-TF-IDF, which ignores the genealogy, largely unaffected.

\begin{figure}[tbp]
  \begin{center}
  \includegraphics[width=.68\textwidth]{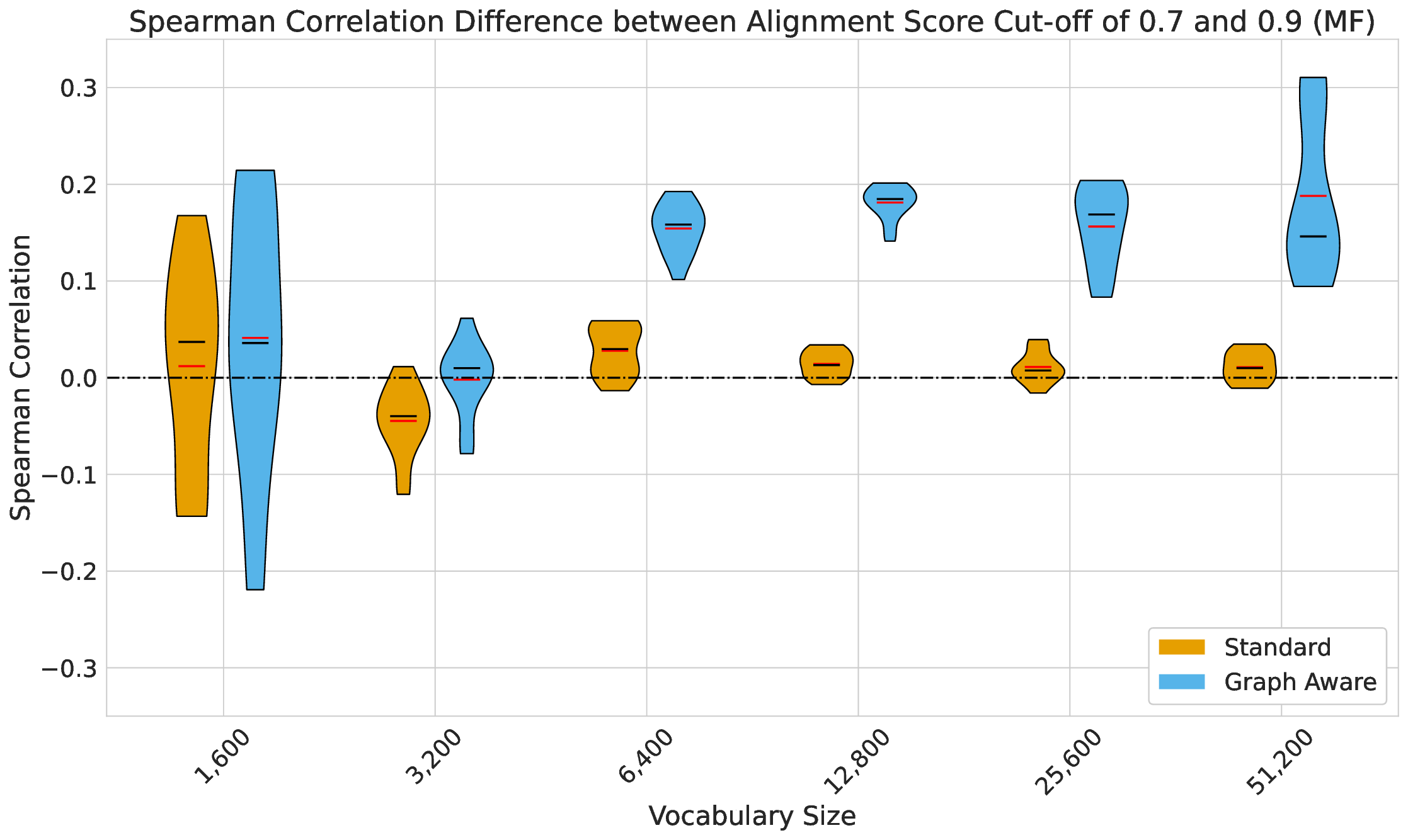}
  \caption{Difference in Spearman correlation between alignment score cut-offs of 0.7 and 0.9 for Molecular Function, shown separately for standard and graph-aware c-TF-IDF. Each observation is a pair of PUMA configurations differing only in the cut-off, with each configuration summarized by its mean over ten held-out GO-term draws; positive values favor the 0.7 cut-off. Black and red bars show median and mean.}
  \label{sup:fig:go_thrs}
  \end{center}
\end{figure}

Figure~\ref{sup:fig:go_thrs} tests this. For every pair of PUMA configurations that differ only in the cut-off, we take the difference in Spearman correlation between the 0.7 and the 0.9 setting, each configuration summarized by its mean over the ten reporting draws; positive values favor the looser cut-off. The distributions are over matched configuration pairs, shown separately for standard and graph-aware c-TF-IDF. Under standard c-TF-IDF, the two cut-offs are indistinguishable at every vocabulary size, with differences centered on zero. Under graph-aware c-TF-IDF, they are indistinguishable only at the two smallest vocabularies; from 6,400 upward the 0.7 cut-off is consistently better and the distribution no longer overlaps zero. The size dependence matches the trend in Figure~\ref{fig:go_topic_all}: at small vocabularies units are common enough that pooling adds little, whereas at large vocabularies units are rare and the broader families admitted at 0.7 supply the evidence that individual units no longer carry.

\section{PUMA Units as Protein Language-Model Tokenizers}
\label{sup:sect:plm}

\noindent This section supports Section~\ref{sec:plm}. Section~\ref{sup:sect:compute} gives the training protocol and compute budget. Section~\ref{sup:sect:robust} shows that the main-text result holds for all three PUMA configurations, and Section~\ref{sup:sect:paired} compares PUMA and BPE task by task. Section~\ref{sup:sect:time} explains the compression and wall-clock numbers. Section~\ref{sup:sect:pertask} gives every task/split result behind Table~\ref{tab:main}.

\subsection{Training protocol and compute budget}
\label{sup:sect:compute}

\noindent \textbf{Arms.} \textbf{PLM-ft} fine-tunes the pretrained encoder end to end on each task. \textbf{Scratch} trains a single-layer transformer encoder (width 320, 8 heads) on each task with no pretraining, so it measures what the tokenizer alone contributes. Both arms use attention1d pooling and a linear projection head. Every configuration is run with at least three random seeds; the reported score is the mean over seeds.

\textbf{Pretraining.} The encoder is the ESM-2 architecture with 12 layers and width 480, trained from random initialization; no released weights are used. Its $\sim$35\,M non-embedding parameters are identical in every run. Only the embedding and output layers scale with the vocabulary, from 0.4\,M parameters (1.1\% of the model) at vocabulary 800 to 24.6\,M (42.3\%) at 51{,}200. The masked-language-model head is tied to the input embedding, as in the reference implementation. Every model was pretrained for one epoch over the full corpus with a batch size of 128 sequences, so all models saw the same protein sequences and took the same number of optimizer steps; what differs between runs is how many tokens each sequence becomes, and therefore the cost of a step.

\textbf{Downstream training.} Batch size is 16 sequences. Runs are capped at 20 epochs with early stopping on the validation split. In both stages, inputs are truncated to 1{,}024 tokens and batches are grouped by length and padded dynamically to the longest member. Truncation is not a confound: averaged over the 18 task/splits, the 1{,}024-token limit discards 0.8\% of residues under AA and 0.1\% under the multi-residue vocabularies.

\textbf{Compute.} All runs used a single NVIDIA H100 GPU. Pretraining the 50 models consumed approximately 500 GPU-hours. The PLM-ft sweep, over all tokenizers, vocabulary sizes, tasks and seeds, consumed approximately 1{,}000 GPU-hours. The Scratch models are single-layer encoders and their training time is small by comparison.

\textbf{Optimizer.} Models with vocabulary sizes of 25{,}600 and 51{,}200 were pretrained with a learning rate of $3\mathrm{e}{-4}$, $\beta_2 = 0.999$, $\epsilon = 1\mathrm{e}{-6}$ and a warmup ratio of 0.05; smaller vocabularies used $4\mathrm{e}{-4}$, $\beta_2 = 0.98$, $\epsilon = 1\mathrm{e}{-8}$ and a warmup ratio of 0.02. The same settings apply to the corresponding PUMA\_PC models. At the two largest sizes the setting used for the smaller vocabularies did not train stably, for BPE as well as for PUMA, and the first setting is the minimal change that restored stable optimization. It was applied identically to every tokenizer at those sizes, so the comparison between tokenizers is unaffected. The instability appears at exactly the sizes where the tied output projection comes to dominate the parameter count, which points to that layer rather than to any particular vocabulary.


\subsection{Robustness across PUMA configurations}
\label{sup:sect:robust}

\noindent Section~\ref{sec:plm} reports PUMA (BLOSUM62, 0.7, 0.05), the configuration that spends the largest share of its vocabulary on mutational variants. Table~\ref{sup:tab:robust} gives the task-balanced mean score of all three configurations, and of their parent-collapsed variants, at every vocabulary size. The three configurations behave alike: all track BPE in PLM-ft, all sit in the same range as BPE in Scratch, and none breaks down at large vocabulary. The choice of configuration therefore does not affect the conclusions of Section~\ref{sec:plm}.

\begin{table}[!htbp]\centering
\caption{\textbf{Task-balanced mean scores for every tokenizer and PUMA configuration.} Splits are averaged within each task and then across the 13 tasks, so tasks with several splits do not dominate.
Deviations are seed standard deviations aggregated the same way. The three PUMA configurations track one another and BPE across the whole vocabulary range in both arms, so the conclusions of the main text do not depend on the configuration reported there. Differences between neighboring cells are frequently smaller than the seed standard deviation (median 0.007 in PLM-ft and 0.018 in Scratch) and should not be read as an ordering.}
\label{sup:tab:robust}\small\setlength{\tabcolsep}{4pt}
\begin{tabular}{l cccccccc}
\toprule
Tokenizer & 33 & .8k & 1.6k & 3.2k & 6.4k & 12.8k & 25.6k & 51.2k\\
\midrule
\multicolumn{9}{l}{\textit{PLM-ft}}\\
AA & 0.671\,{\tiny$\pm$.009} &   &   &   &   &   &   &   \\
BPE & & 0.654\,{\tiny$\pm$.009} & 0.656\,{\tiny$\pm$.009} & 0.643\,{\tiny$\pm$.011} & 0.642\,{\tiny$\pm$.011} & 0.630\,{\tiny$\pm$.008} & 0.635\,{\tiny$\pm$.011} & 0.623\,{\tiny$\pm$.031} \\
PUMA (B62, 0.7) & & 0.651\,{\tiny$\pm$.007} & 0.643\,{\tiny$\pm$.015} & 0.636\,{\tiny$\pm$.017} & 0.637\,{\tiny$\pm$.012} & 0.646\,{\tiny$\pm$.009} & 0.636\,{\tiny$\pm$.011} & 0.639\,{\tiny$\pm$.010} \\
PUMA (B62, 0.8) & & 0.656\,{\tiny$\pm$.013} & 0.646\,{\tiny$\pm$.016} & 0.647\,{\tiny$\pm$.008} & 0.643\,{\tiny$\pm$.010} & 0.633\,{\tiny$\pm$.012} & 0.642\,{\tiny$\pm$.010} & 0.627\,{\tiny$\pm$.022} \\
PUMA (PAM70, 0.7) & & 0.653\,{\tiny$\pm$.010} & 0.648\,{\tiny$\pm$.018} & 0.653\,{\tiny$\pm$.012} & 0.648\,{\tiny$\pm$.012} & 0.632\,{\tiny$\pm$.011} & 0.636\,{\tiny$\pm$.011} & 0.637\,{\tiny$\pm$.012} \\
PUMA\_PC (B62, 0.7) & & 0.649\,{\tiny$\pm$.011} & 0.644\,{\tiny$\pm$.013} & 0.636\,{\tiny$\pm$.009} & 0.632\,{\tiny$\pm$.010} & 0.623\,{\tiny$\pm$.009} & 0.619\,{\tiny$\pm$.014} & 0.619\,{\tiny$\pm$.009} \\
PUMA\_PC (B62, 0.8) & & 0.652\,{\tiny$\pm$.011} & 0.649\,{\tiny$\pm$.016} & 0.645\,{\tiny$\pm$.009} & 0.638\,{\tiny$\pm$.010} & 0.632\,{\tiny$\pm$.008} & 0.615\,{\tiny$\pm$.019} & 0.630\,{\tiny$\pm$.009} \\
PUMA\_PC (PAM70, 0.7) & & 0.651\,{\tiny$\pm$.012} & 0.636\,{\tiny$\pm$.012} & 0.625\,{\tiny$\pm$.010} & 0.627\,{\tiny$\pm$.008} & 0.627\,{\tiny$\pm$.009} & 0.626\,{\tiny$\pm$.012} & 0.619\,{\tiny$\pm$.015} \\
\midrule
\multicolumn{9}{l}{\textit{Scratch}}\\
AA & 0.406\,{\tiny$\pm$.033} &   &   &   &   &   &   &   \\
BPE & & 0.438\,{\tiny$\pm$.027} & 0.431\,{\tiny$\pm$.039} & 0.432\,{\tiny$\pm$.036} & 0.447\,{\tiny$\pm$.033} & 0.427\,{\tiny$\pm$.051} & 0.444\,{\tiny$\pm$.030} & 0.448\,{\tiny$\pm$.014} \\
PUMA (B62, 0.7) & & 0.409\,{\tiny$\pm$.031} & 0.448\,{\tiny$\pm$.031} & 0.427\,{\tiny$\pm$.023} & 0.443\,{\tiny$\pm$.029} & 0.452\,{\tiny$\pm$.026} & 0.449\,{\tiny$\pm$.024} & 0.455\,{\tiny$\pm$.030} \\
PUMA (B62, 0.8) & & 0.430\,{\tiny$\pm$.039} & 0.436\,{\tiny$\pm$.042} & 0.431\,{\tiny$\pm$.041} & 0.453\,{\tiny$\pm$.044} & 0.448\,{\tiny$\pm$.029} & 0.443\,{\tiny$\pm$.039} & 0.455\,{\tiny$\pm$.035} \\
PUMA (PAM70, 0.7) & & 0.377\,{\tiny$\pm$.038} & 0.434\,{\tiny$\pm$.044} & 0.435\,{\tiny$\pm$.049} & 0.451\,{\tiny$\pm$.023} & 0.451\,{\tiny$\pm$.022} & 0.433\,{\tiny$\pm$.050} & 0.448\,{\tiny$\pm$.040} \\
PUMA\_PC (B62, 0.7) & & 0.422\,{\tiny$\pm$.033} & 0.449\,{\tiny$\pm$.027} & 0.421\,{\tiny$\pm$.028} & 0.440\,{\tiny$\pm$.029} & 0.432\,{\tiny$\pm$.044} & 0.413\,{\tiny$\pm$.043} & 0.460\,{\tiny$\pm$.025} \\
PUMA\_PC (B62, 0.8) & & 0.418\,{\tiny$\pm$.039} & 0.435\,{\tiny$\pm$.034} & 0.440\,{\tiny$\pm$.046} & 0.447\,{\tiny$\pm$.041} & 0.462\,{\tiny$\pm$.032} & 0.433\,{\tiny$\pm$.038} & 0.441\,{\tiny$\pm$.031} \\
PUMA\_PC (PAM70, 0.7) & & 0.429\,{\tiny$\pm$.036} & 0.403\,{\tiny$\pm$.043} & 0.416\,{\tiny$\pm$.044} & 0.429\,{\tiny$\pm$.038} & 0.455\,{\tiny$\pm$.035} & 0.458\,{\tiny$\pm$.025} & 0.455\,{\tiny$\pm$.032} \\
\bottomrule
\end{tabular}
\end{table}

\subsection{Paired comparison of PUMA and BPE}
\label{sup:sect:paired}

\noindent Table~\ref{sup:tab:paired} supports the claim in Section~\ref{sec:plm} that PUMA and BPE perform comparably. Each cell is the difference between PUMA and BPE on one task at one vocabulary size, computed from seed means after averaging over splits within a task. For each vocabulary size and arm, a two-sided Wilcoxon signed-rank test checks whether the 13 paired differences are centered on zero. Thirteen of the 14 tests are not significant: $p$ ranges from 0.09 to 1.00 in PLM-ft and from 0.22 to 1.00 in Scratch. The exception is PLM-ft at 12{,}800, where PUMA scores higher than BPE on 11 of the 13 tasks ($p = 0.002$; $p = 0.02$ after Bonferroni correction over the 14 tests); this advantage does not persist at 25{,}600 or 51{,}200. In the other 13 comparisons PUMA scores higher on between 4 and 9 of the 13 tasks, and the median difference is within 0.01 of zero in every comparison except PLM-ft at 12{,}800 ($+0.011$).

In PLM-ft the differences are small, with 78 of the 91 cells within 0.02 of zero. They are larger in Scratch, where 55 of the 91 cells are within 0.02, which matches the larger seed variance of the single-layer encoder (Table~\ref{sup:tab:sc}); they still fall in both directions. The largest differences in both arms are on AAV, where the sign changes repeatedly across vocabulary sizes and seed standard deviations reach 0.21. We therefore attribute them to the instability of this task rather than to the tokenizer. The only task with a consistent direction is GB1 in PLM-ft, where PUMA scores below BPE at the four smallest vocabulary sizes, by 0.02 to 0.05, and level with it from 12{,}800 upward. GB1 also has large seed variance, up to 0.09 for these two tokenizers in PLM-ft (Table~\ref{sup:tab:ft}), and this single task does not change the result across tasks.

\begin{table}[!htbp]\centering
\caption{\textbf{Per-task difference between PUMA and BPE.} Each cell is PUMA $-$ BPE for one task and vocabulary size (seed means, averaged over splits within a task); positive values favor PUMA. PUMA: BLOSUM62, 0.7, 0.05. PLM-ft: pretrained encoder, fully fine-tuned. Scratch: single-layer encoder trained on each task without pretraining.}
\label{sup:tab:paired}\small\setlength{\tabcolsep}{3pt}
\begin{tabular}{l *{7}{c}@{\hspace{8pt}}*{7}{c}}
\toprule
& \multicolumn{7}{c}{\textbf{PLM-ft}} & \multicolumn{7}{c}{\textbf{Scratch}}\\
\cmidrule(lr){2-8}\cmidrule(lr){9-15}
Task & .8k & 1.6k & 3.2k & 6.4k & 12.8k & 25.6k & 51.2k & .8k & 1.6k & 3.2k & 6.4k & 12.8k & 25.6k & 51.2k\\
\midrule
\addlinespace[1.5pt]\multicolumn{15}{l}{\textit{Variant effects}}\\[0.5pt]
Fluorescence & .00 & .00 & .00 & .00 & .00 & .00 & .00 & $-$.01 & $-$.06 & $-$.02 & +.04 & +.04 & $-$.02 & $-$.03\\
AAV & +.02 & $-$.13 & $-$.06 & $-$.03 & +.01 & $-$.01 & +.14 & $-$.26 & +.25 & $-$.01 & $-$.12 & +.13 & +.02 & +.10\\
GB1 & $-$.02 & $-$.05 & $-$.04 & $-$.03 & .00 & .00 & .00 & $-$.04 & $-$.02 & +.02 & $-$.08 & $-$.02 & $-$.03 & $-$.01\\
Stability & .00 & +.02 & +.01 & .00 & +.06 & $-$.01 & $-$.02 & +.02 & +.03 & $-$.05 & $-$.03 & +.13 & $-$.02 & $-$.04\\
SolMut CS & .00 & .00 & .00 & .00 & +.03 & +.02 & +.05 & $-$.07 & +.02 & $-$.03 & +.08 & +.08 & +.02 & +.04\\
SolMut BLAT & $-$.04 & $-$.01 & $-$.01 & +.01 & +.03 & +.01 & .00 & $-$.01 & $-$.03 & $-$.03 & .00 & $-$.05 & +.06 & +.06\\
SolMut LGK & .00 & .00 & .00 & +.01 & +.01 & .00 & +.01 & +.06 & $-$.01 & +.02 & +.01 & $-$.05 & +.02 & +.01\\
\addlinespace[1.5pt]\multicolumn{15}{l}{\textit{Protein properties}}\\[0.5pt]
Meltome & +.01 & .00 & .00 & $-$.01 & +.03 & +.01 & +.01 & .00 & +.02 & $-$.01 & $-$.03 & $-$.01 & .00 & .00\\
DeepSol & $-$.01 & $-$.01 & .00 & .00 & +.01 & .00 & $-$.01 & $-$.07 & +.04 & +.04 & +.06 & +.08 & +.01 & +.01\\
DeepLoc-2 & .00 & .00 & .00 & $-$.01 & +.01 & $-$.01 & .00 & .00 & .00 & .00 & .00 & .00 & .00 & .00\\
DeepLoc bin. & .00 & $-$.01 & .00 & +.01 & .00 & .00 & .00 & $-$.01 & .00 & .00 & $-$.01 & $-$.01 & $-$.01 & $-$.03\\
\addlinespace[1.5pt]\multicolumn{15}{l}{\textit{Homology and interaction}}\\[0.5pt]
Remote homology & .00 & .00 & .00 & .00 & +.02 & $-$.01 & +.01 & +.01 & $-$.01 & +.01 & +.02 & +.01 & .00 & .00\\
PPI SHS27k & .00 & +.01 & +.01 & .00 & .00 & .00 & .00 & .00 & .00 & +.01 & $-$.01 & $-$.01 & +.01 & $-$.01\\
\bottomrule
\end{tabular}
\end{table}

\subsection{Compression and wall-clock cost}
\label{sup:sect:time}

\noindent Two forces act on runtime, in opposite directions. Compression shortens the input: the mean number of residues per token, $\bar{k}$, rises from 1.00 for AA to about 2.9--3.0 at vocabulary 51{,}200 (Table~\ref{sup:tab:ft_time}), and every encoder layer becomes cheaper in proportion. Vocabulary size acts against this, but only during pretraining, where the masked-language-model objective projects onto the full vocabulary at every masked position; at width 480 this layer alone is 42.3\% of the model's parameters at 51{,}200 against 1.1\% at 800. The net effect is the U-shape in Table~\ref{sup:tab:pretrain_time} for BPE and full PUMA.

Parent collapse separates the two forces. The sequence is segmented by the full PUMA vocabulary, so $\bar{k}$ is unchanged, but the model predicts over the collapsed vocabulary (Table~\ref{sup:tab:conv}). PUMA\_PC therefore receives the compression benefit without the projection penalty, and it is the only vocabulary whose pretraining time falls monotonically across the whole range.

Fine-tuning has no vocabulary-sized output layer, since the task head is a low-dimensional projection, so its cost is set by sequence length alone and all downstream speedups in Table~\ref{sup:tab:ft_time} increase with vocabulary. Downstream runs stop early, so these times are the cost of bringing a task to convergence rather than pure throughput, and a tokenizer that converges in fewer epochs is credited for it. The pattern nonetheless suggests throughput dominates: across the 21 tokenizer/vocabulary cells, the fine-tuning speedup tracks $\bar{k}$ almost exactly (Spearman $\rho = 0.97$), as a per-step cost proportional to sequence length predicts. Pretraining entries are single runs and downstream entries are seed means, so differences of a few percent are within run-to-run noise.

\begin{table}[h]\centering
\caption{\textbf{Vocabulary contraction under parent collapse.} A sequence is tokenized with the original PUMA tokenizer and each token is then remapped to the mutational parent of its family; singletons are unchanged. Because most PUMA tokens are rare mutational children of a shared parent, the trained embedding table shrinks sharply, and increasingly so with vocabulary size. The contraction is strongest for the configurations that spend most of their budget on mutational variants.}
\label{sup:tab:conv}\small\setlength{\tabcolsep}{5pt}
\begin{tabular}{r rr rr rr}
\toprule
& \multicolumn{2}{c}{BLOSUM62, 0.7} & \multicolumn{2}{c}{BLOSUM62, 0.8} & \multicolumn{2}{c}{PAM70, 0.7}\\
\cmidrule(lr){2-3}\cmidrule(lr){4-5}\cmidrule(lr){6-7}
PUMA vocab size & PUMA\_PC & Reduction & PUMA\_PC & Reduction & PUMA\_PC & Reduction\\
\midrule
800 & 389 & 2.1$\times$ & 531 & 1.5$\times$ & 404 & 2.0$\times$ \\
1,600 & 513 & 3.1$\times$ & 896 & 1.8$\times$ & 579 & 2.8$\times$ \\
3,200 & 730 & 4.4$\times$ & 1,446 & 2.2$\times$ & 873 & 3.7$\times$ \\
6,400 & 1,044 & 6.1$\times$ & 2,168 & 3.0$\times$ & 1,315 & 4.9$\times$ \\
12,800 & 1,481 & 8.6$\times$ & 3,375 & 3.8$\times$ & 2,051 & 6.2$\times$ \\
25,600 & 2,314 & 11.1$\times$ & 5,732 & 4.5$\times$ & 3,408 & 7.5$\times$ \\
51,200 & 3,977 & 12.9$\times$ & 10,342 & 5.0$\times$ & 6,058 & 8.5$\times$ \\
\bottomrule
\end{tabular}
\end{table}

\begin{table}[htbp]\centering
\caption{\textbf{Pretraining wall-clock cost.} Hours for one epoch of masked-language-model pretraining over UniRef50 (2024-06), and the corresponding speedup relative to the single-residue baseline. PUMA and PUMA\_PC rows are averaged over the configurations of Table~\ref{sup:tab:robust} with a recorded runtime (one to three per vocabulary size), whose individual times differ by at most 6\% at any vocabulary size. Every multi-residue tokenizer is faster than AA, but BPE and full PUMA reach a minimum near 3{,}200--6{,}400 and slow again beyond it, because the masked-language-model objective projects onto the full vocabulary at every masked position. Parent collapse leaves the segmentation unchanged while shrinking that projection, and is the only vocabulary whose pretraining time falls monotonically across the whole range.}
\label{sup:tab:pretrain_time}\small\setlength{\tabcolsep}{5pt}
\begin{tabular}{l c ccccccc}
\toprule
Tokenizer & 33 & .8k & 1.6k & 3.2k & 6.4k & 12.8k & 25.6k & 51.2k\\
\midrule
\multicolumn{9}{l}{\textit{Wall-clock hours}}\\
AA & 18.18 & & & & & & & \\
BPE & & 11.82 & 11.12 & 10.78 & 10.77 & 11.11 & 15.56 & 16.97\\
PUMA & & 11.91 & 11.41 & 11.18 & 11.08 & 11.31 & 15.45 & 16.85\\
PUMA\_PC & & 11.84 & 11.22 & 10.86 & 10.55 & 10.24 & 10.10 & 9.94\\
\midrule
\multicolumn{9}{l}{\textit{Speedup vs.\ AA}}\\
AA & 1.00$\times$ & & & & & & & \\
BPE & & 1.54$\times$ & 1.64$\times$ & 1.69$\times$ & 1.69$\times$ & 1.64$\times$ & 1.17$\times$ & 1.07$\times$\\
PUMA & & 1.53$\times$ & 1.59$\times$ & 1.63$\times$ & 1.64$\times$ & 1.61$\times$ & 1.18$\times$ & 1.08$\times$\\
PUMA\_PC & & 1.54$\times$ & 1.62$\times$ & 1.67$\times$ & 1.72$\times$ & 1.78$\times$ & 1.80$\times$ & 1.83$\times$\\
\bottomrule
\end{tabular}
\end{table}

\begin{table}[htbp]\centering
\caption{\textbf{Compression and downstream wall-clock cost.} \textit{Top block:} mean compression $\bar{k}$, residues encoded per token, averaged over the 18 downstream task/splits. Parent collapse relabels an unchanged segmentation, so PUMA and PUMA\_PC compress identically and differ only in the size of the embedding and output layers. \textit{Bottom block:} fine-tuning speedup relative to AA, computed per task/split as the ratio of the seed-mean AA runtime to the seed-mean tokenizer runtime and aggregated as a geometric mean over the 18 task/splits. PUMA and PUMA\_PC rows are averaged over the three configurations. Fine-tuning carries no vocabulary-sized output layer, so its cost is set by sequence length alone: downstream speedups rise with vocabulary and never invert as they do in pretraining.}
\label{sup:tab:ft_time}\small\setlength{\tabcolsep}{5pt}
\begin{tabular}{l c ccccccc}
\toprule
Tokenizer & 33 & .8k & 1.6k & 3.2k & 6.4k & 12.8k & 25.6k & 51.2k\\
\midrule
\multicolumn{9}{l}{\textit{Compression} $\bar{k}$ \textit{(residues per token)}}\\
AA & 1.00 & & & & & & & \\
BPE & & 1.93 & 2.13 & 2.30 & 2.45 & 2.62 & 2.80 & 3.02\\
PUMA & & 1.89 & 2.07 & 2.24 & 2.39 & 2.55 & 2.74 & 2.94\\
PUMA\_PC & & 1.89 & 2.07 & 2.24 & 2.39 & 2.55 & 2.74 & 2.94\\
\midrule
\multicolumn{9}{l}{\textit{PLM-ft --- speedup vs.\ AA}}\\
AA & 1.00$\times$ & & & & & & & \\
BPE & & 1.53$\times$ & 1.55$\times$ & 1.54$\times$ & 1.68$\times$ & 1.70$\times$ & 1.74$\times$ & 1.73$\times$\\
PUMA & & 1.40$\times$ & 1.44$\times$ & 1.55$\times$ & 1.61$\times$ & 1.65$\times$ & 1.69$\times$ & 1.81$\times$\\
PUMA\_PC & & 1.43$\times$ & 1.50$\times$ & 1.57$\times$ & 1.62$\times$ & 1.63$\times$ & 1.70$\times$ & 1.74$\times$\\
\bottomrule
\end{tabular}
\end{table}

\subsection{Complete per-task results}
\label{sup:sect:pertask}

\noindent Table~\ref{sup:tab:tasks} gives details of each task; Tables~\ref{sup:tab:ft} and~\ref{sup:tab:sc} give all 18 task/splits for the BLOSUM62, 0.7, 0.05 configuration in both arms, at all seven vocabulary sizes.
Table~\ref{tab:main} reports one split per task from them: two-vs-rest for GB1, the mixed split for Meltome, and the fold holdout for remote homology.
Bold entries are maxima over 22 columns whose seed standard deviations are often as large as the differences between them, and should not be read as rankings.

\begin{table*}[!t]
\caption{\textbf{Downstream tasks.} Task definitions, splits and metrics follow the PETA benchmark \citep{tan2024peta}; we use 13 of its tasks, evaluated on 18 task/splits. The three SolMut tasks are listed together because they share a definition. $^{\dagger}$Split reported in Table~\ref{tab:main} for tasks with more than one split. $^{\ast}$Multi-label accuracy, averaged over the ten compartments. All metrics are higher-is-better.}
\label{sup:tab:tasks}
\centering
\footnotesize
\setlength{\tabcolsep}{4pt}
\begin{tabular}{@{}>{\raggedright\arraybackslash}p{0.1\textwidth} >{\raggedright\arraybackslash}p{0.11\textwidth} >{\raggedright\arraybackslash}p{0.41\textwidth} >{\raggedright\arraybackslash}p{0.1\textwidth} c >{\raggedright\arraybackslash}p{0.14\textwidth}@{}}
\toprule
Group & Task & Description & Metric & Splits & Split names \\
\midrule
Variant effects & Fluorescence & Log-fluorescence of GFP mutants; trained on low-order mutants and tested on higher-order ones. & Spearman $\rho$ & 1 & default \\
 & AAV & Fitness of AAV2 capsid (VP1) variants mutated in positions 561--588; trained on variants with at most two mutations and tested on the rest. & Spearman $\rho$ & 1 & two-vs-many \\
 & GB1 & Binding fitness of GB1 variants at four epistatic positions. Splits train on variants with at most two or three mutations, or on low-fitness variants, and test on the rest. & Spearman $\rho$ & 3 & two-vs-rest$^{\dagger}$, three-vs-rest, low-vs-high \\
 & Stability & Stability score of designed mini-proteins; tested on single-mutant neighbors of the most stable designs. & Spearman $\rho$ & 1 & default \\
 & SolMut CS / BLAT / LGK & Change in solubility upon mutation (SoluProtMutDB) for chalcone synthase, TEM-1 $\beta$-lactamase and levoglucosan kinase; one task per protein, random 8:1:1 split. & Spearman $\rho$ & 1 each & default (each) \\
\addlinespace
Protein properties & Meltome & Thermostability from mass-spectrometry melting curves of proteins from 13 species. & Spearman $\rho$ & 2 & mixed split$^{\dagger}$, human \\
 & DeepSol & Binary solubility (soluble or insoluble); training sequences with $\geq$30\% identity to the test set are removed. & Accuracy & 1 & default \\
 & DeepLoc-2 & Multi-label subcellular localization over ten compartments; tested on a held-out Swiss-Prot partition. & Accuracy$^{\ast}$ & 1 & test \\
 & DeepLoc binary & Membrane-bound versus soluble proteins from DeepLoc-1. & Accuracy & 1 & default \\
\addlinespace
Homology and interaction & Remote homology & Fold classification into 1,195 SCOP folds. In the family, superfamily and fold holdouts, the closest training homolog of a test protein shares, respectively, its family, only its superfamily, or only its fold. & Accuracy & 3 & fold$^{\dagger}$, superfamily, family holdout \\
 & PPI SHS27k & Interaction type (seven classes) for pairs of human proteins from STRING; random 8:1:1 split. & Accuracy & 1 & default \\
\bottomrule
\end{tabular}
\end{table*}

\begin{sidewaystable}[p]\centering
\caption{\textbf{PLM-ft} --- pretrained encoder, full fine-tuning; every task/split. PUMA and PUMA\_PC are the BLOSUM62,\,0.7,\,0.05 configuration. Mean over seeds with the seed standard deviation beneath; higher is better; 
\textbf{bold} marks the best value in each row, including ties. AA leads on most splits, with the remote homology, GB1 and AAV splits carrying most of the gap. PUMA tracks BPE across the whole vocabulary range.}
\label{sup:tab:ft}
\tiny\setlength{\tabcolsep}{3pt}
\begin{tabular}{l c ccccccc ccccccc ccccccc}
\toprule
& AA & \multicolumn{7}{c}{BPE} & \multicolumn{7}{c}{PUMA} & \multicolumn{7}{c}{PUMA\_PC}\\
\cmidrule(lr){3-9}\cmidrule(lr){10-16}\cmidrule(lr){17-23}
Task/split & 33 & .8k & 1.6k & 3.2k & 6.4k & 12.8k & 25.6k & 51.2k & .8k & 1.6k & 3.2k & 6.4k & 12.8k & 25.6k & 51.2k & .8k & 1.6k & 3.2k & 6.4k & 12.8k & 25.6k & 51.2k\\
\midrule
fluorescence/default & 0.677 & 0.678 & 0.678 & 0.677 & 0.675 & 0.679 & 0.674 & 0.677 & 0.677 & \textbf{0.680} & 0.679 & 0.676 & 0.676 & 0.675 & 0.679 & 0.678 & 0.674 & 0.671 & 0.668 & 0.663 & 0.662 & 0.662 \\
 & {\tiny$\pm$.003} & {\tiny$\pm$.003} & {\tiny$\pm$.002} & {\tiny$\pm$.001} & {\tiny$\pm$.001} & {\tiny$\pm$.002} & {\tiny$\pm$.002} & {\tiny$\pm$.002} & {\tiny$\pm$.003} & {\tiny$\pm$.002} & {\tiny$\pm$.002} & {\tiny$\pm$.002} & {\tiny$\pm$.001} & {\tiny$\pm$.001} & {\tiny$\pm$.001} & {\tiny$\pm$.001} & {\tiny$\pm$.002} & {\tiny$\pm$.001} & {\tiny$\pm$.001} & {\tiny$\pm$.002} & {\tiny$\pm$.002} & {\tiny$\pm$.001} \\[1pt]
stability/default & \textbf{0.775} & 0.737 & 0.734 & 0.729 & 0.724 & 0.690 & 0.707 & 0.739 & 0.735 & 0.757 & 0.739 & 0.728 & 0.750 & 0.702 & 0.720 & 0.754 & 0.712 & 0.712 & 0.733 & 0.686 & 0.699 & 0.710 \\
 & {\tiny$\pm$.013} & {\tiny$\pm$.016} & {\tiny$\pm$.009} & {\tiny$\pm$.030} & {\tiny$\pm$.024} & {\tiny$\pm$.005} & {\tiny$\pm$.016} & {\tiny$\pm$.029} & {\tiny$\pm$.012} & {\tiny$\pm$.007} & {\tiny$\pm$.024} & {\tiny$\pm$.023} & {\tiny$\pm$.031} & {\tiny$\pm$.035} & {\tiny$\pm$.035} & {\tiny$\pm$.023} & {\tiny$\pm$.029} & {\tiny$\pm$.019} & {\tiny$\pm$.040} & {\tiny$\pm$.002} & {\tiny$\pm$.023} & {\tiny$\pm$.017} \\[1pt]
aav/two\_vs\_many & 0.811 & 0.813 & 0.811 & 0.761 & 0.789 & 0.770 & 0.792 & 0.642 & \textbf{0.830} & 0.686 & 0.698 & 0.758 & 0.781 & 0.783 & 0.783 & 0.813 & 0.752 & 0.780 & 0.761 & 0.733 & 0.690 & 0.710 \\
 & {\tiny$\pm$.010} & {\tiny$\pm$.006} & {\tiny$\pm$.005} & {\tiny$\pm$.028} & {\tiny$\pm$.020} & {\tiny$\pm$.007} & {\tiny$\pm$.009} & {\tiny$\pm$.195} & {\tiny$\pm$.005} & {\tiny$\pm$.117} & {\tiny$\pm$.116} & {\tiny$\pm$.003} & {\tiny$\pm$.013} & {\tiny$\pm$.010} & {\tiny$\pm$.007} & {\tiny$\pm$.008} & {\tiny$\pm$.015} & {\tiny$\pm$.015} & {\tiny$\pm$.012} & {\tiny$\pm$.013} & {\tiny$\pm$.048} & {\tiny$\pm$.009} \\[1pt]
gb1/two\_vs\_rest & \textbf{0.585} & 0.391 & 0.534 & 0.515 & 0.513 & 0.500 & 0.509 & 0.521 & 0.403 & 0.491 & 0.487 & 0.501 & 0.438 & 0.465 & 0.500 & 0.466 & 0.543 & 0.423 & 0.494 & 0.477 & 0.493 & 0.423 \\
 & {\tiny$\pm$.034} & {\tiny$\pm$.009} & {\tiny$\pm$.036} & {\tiny$\pm$.004} & {\tiny$\pm$.080} & {\tiny$\pm$.037} & {\tiny$\pm$.046} & {\tiny$\pm$.013} & {\tiny$\pm$.034} & {\tiny$\pm$.013} & {\tiny$\pm$.050} & {\tiny$\pm$.055} & {\tiny$\pm$.040} & {\tiny$\pm$.079} & {\tiny$\pm$.022} & {\tiny$\pm$.138} & {\tiny$\pm$.082} & {\tiny$\pm$.083} & {\tiny$\pm$.020} & {\tiny$\pm$.079} & {\tiny$\pm$.072} & {\tiny$\pm$.032} \\[1pt]
gb1/three\_vs\_rest & \textbf{0.866} & 0.836 & 0.855 & 0.839 & 0.840 & 0.839 & 0.847 & 0.839 & 0.850 & 0.843 & 0.837 & 0.831 & 0.832 & 0.833 & 0.836 & 0.856 & 0.842 & 0.840 & 0.835 & 0.824 & 0.833 & 0.817 \\
 & {\tiny$\pm$.003} & {\tiny$\pm$.003} & {\tiny$\pm$.002} & {\tiny$\pm$.005} & {\tiny$\pm$.003} & {\tiny$\pm$.003} & {\tiny$\pm$.005} & {\tiny$\pm$.006} & {\tiny$\pm$.002} & {\tiny$\pm$.004} & {\tiny$\pm$.003} & {\tiny$\pm$.005} & {\tiny$\pm$.006} & {\tiny$\pm$.005} & {\tiny$\pm$.009} & {\tiny$\pm$.005} & {\tiny$\pm$.005} & {\tiny$\pm$.002} & {\tiny$\pm$.003} & {\tiny$\pm$.008} & {\tiny$\pm$.001} & {\tiny$\pm$.005} \\[1pt]
gb1/low\_vs\_high & \textbf{0.410} & 0.386 & 0.371 & 0.363 & 0.338 & 0.306 & 0.248 & 0.284 & 0.299 & 0.291 & 0.285 & 0.261 & 0.382 & 0.313 & 0.322 & 0.345 & 0.357 & 0.360 & 0.346 & 0.297 & 0.320 & 0.322 \\
 & {\tiny$\pm$.075} & {\tiny$\pm$.020} & {\tiny$\pm$.020} & {\tiny$\pm$.061} & {\tiny$\pm$.026} & {\tiny$\pm$.032} & {\tiny$\pm$.047} & {\tiny$\pm$.041} & {\tiny$\pm$.026} & {\tiny$\pm$.018} & {\tiny$\pm$.014} & {\tiny$\pm$.086} & {\tiny$\pm$.002} & {\tiny$\pm$.031} & {\tiny$\pm$.027} & {\tiny$\pm$.023} & {\tiny$\pm$.043} & {\tiny$\pm$.006} & {\tiny$\pm$.007} & {\tiny$\pm$.034} & {\tiny$\pm$.012} & {\tiny$\pm$.034} \\[1pt]
solmut\_cs/default & 0.492 & 0.514 & 0.505 & 0.516 & 0.483 & 0.474 & 0.478 & 0.448 & 0.514 & 0.509 & \textbf{0.519} & 0.478 & 0.504 & 0.503 & 0.496 & 0.484 & 0.486 & 0.481 & 0.425 & 0.475 & 0.408 & 0.456 \\
 & {\tiny$\pm$.015} & {\tiny$\pm$.007} & {\tiny$\pm$.017} & {\tiny$\pm$.001} & {\tiny$\pm$.016} & {\tiny$\pm$.007} & {\tiny$\pm$.012} & {\tiny$\pm$.098} & {\tiny$\pm$.005} & {\tiny$\pm$.005} & {\tiny$\pm$.002} & {\tiny$\pm$.014} & {\tiny$\pm$.013} & {\tiny$\pm$.012} & {\tiny$\pm$.005} & {\tiny$\pm$.006} & {\tiny$\pm$.031} & {\tiny$\pm$.007} & {\tiny$\pm$.015} & {\tiny$\pm$.008} & {\tiny$\pm$.032} & {\tiny$\pm$.005} \\[1pt]
solmut\_blat/default & 0.517 & 0.535 & \textbf{0.545} & 0.486 & 0.512 & 0.464 & 0.483 & 0.481 & 0.496 & 0.531 & 0.477 & 0.517 & 0.489 & 0.490 & 0.484 & 0.508 & 0.515 & 0.472 & 0.486 & 0.467 & 0.484 & 0.461 \\
 & {\tiny$\pm$.008} & {\tiny$\pm$.029} & {\tiny$\pm$.015} & {\tiny$\pm$.004} & {\tiny$\pm$.008} & {\tiny$\pm$.017} & {\tiny$\pm$.014} & {\tiny$\pm$.013} & {\tiny$\pm$.013} & {\tiny$\pm$.016} & {\tiny$\pm$.008} & {\tiny$\pm$.008} & {\tiny$\pm$.013} & {\tiny$\pm$.007} & {\tiny$\pm$.020} & {\tiny$\pm$.008} & {\tiny$\pm$.014} & {\tiny$\pm$.015} & {\tiny$\pm$.002} & {\tiny$\pm$.004} & {\tiny$\pm$.004} & {\tiny$\pm$.012} \\[1pt]
solmut\_lgk/default & \textbf{0.585} & 0.554 & 0.551 & 0.534 & 0.522 & 0.509 & 0.520 & 0.510 & 0.555 & 0.550 & 0.533 & 0.528 & 0.524 & 0.520 & 0.523 & 0.543 & 0.551 & 0.536 & 0.543 & 0.540 & 0.538 & 0.524 \\
 & {\tiny$\pm$.004} & {\tiny$\pm$.005} & {\tiny$\pm$.010} & {\tiny$\pm$.007} & {\tiny$\pm$.007} & {\tiny$\pm$.001} & {\tiny$\pm$.003} & {\tiny$\pm$.005} & {\tiny$\pm$.010} & {\tiny$\pm$.006} & {\tiny$\pm$.004} & {\tiny$\pm$.002} & {\tiny$\pm$.007} & {\tiny$\pm$.004} & {\tiny$\pm$.005} & {\tiny$\pm$.003} & {\tiny$\pm$.006} & {\tiny$\pm$.006} & {\tiny$\pm$.006} & {\tiny$\pm$.003} & {\tiny$\pm$.006} & {\tiny$\pm$.010} \\[1pt]
meltome/mixed\_split & \textbf{0.612} & 0.596 & 0.596 & 0.586 & 0.598 & 0.573 & 0.585 & 0.581 & \textbf{0.612} & 0.593 & 0.587 & 0.584 & 0.596 & 0.586 & 0.591 & 0.587 & 0.588 & 0.587 & 0.563 & 0.538 & 0.543 & 0.540 \\
 & {\tiny$\pm$.005} & {\tiny$\pm$.007} & {\tiny$\pm$.008} & {\tiny$\pm$.013} & {\tiny$\pm$.003} & {\tiny$\pm$.010} & {\tiny$\pm$.006} & {\tiny$\pm$.009} & {\tiny$\pm$.002} & {\tiny$\pm$.006} & {\tiny$\pm$.006} & {\tiny$\pm$.011} & {\tiny$\pm$.006} & {\tiny$\pm$.005} & {\tiny$\pm$.010} & {\tiny$\pm$.011} & {\tiny$\pm$.006} & {\tiny$\pm$.004} & {\tiny$\pm$.012} & {\tiny$\pm$.012} & {\tiny$\pm$.008} & {\tiny$\pm$.006} \\[1pt]
meltome/human & 0.691 & 0.701 & 0.699 & 0.691 & 0.693 & 0.674 & 0.670 & 0.667 & 0.701 & 0.697 & 0.691 & 0.691 & \textbf{0.703} & 0.683 & 0.679 & 0.690 & 0.692 & 0.695 & 0.678 & 0.683 & 0.684 & 0.677 \\
 & {\tiny$\pm$.001} & {\tiny$\pm$.004} & {\tiny$\pm$.006} & {\tiny$\pm$.013} & {\tiny$\pm$.003} & {\tiny$\pm$.009} & {\tiny$\pm$.022} & {\tiny$\pm$.003} & {\tiny$\pm$.003} & {\tiny$\pm$.007} & {\tiny$\pm$.004} & {\tiny$\pm$.008} & {\tiny$\pm$.002} & {\tiny$\pm$.004} & {\tiny$\pm$.002} & {\tiny$\pm$.012} & {\tiny$\pm$.016} & {\tiny$\pm$.004} & {\tiny$\pm$.024} & {\tiny$\pm$.010} & {\tiny$\pm$.005} & {\tiny$\pm$.009} \\[1pt]
deepsol/default & 0.730 & 0.732 & 0.732 & 0.727 & 0.726 & 0.724 & 0.735 & 0.730 & 0.725 & 0.723 & 0.724 & 0.724 & 0.731 & 0.733 & 0.724 & \textbf{0.737} & 0.730 & 0.732 & 0.736 & \textbf{0.737} & 0.715 & 0.721 \\
 & {\tiny$\pm$.010} & {\tiny$\pm$.010} & {\tiny$\pm$.006} & {\tiny$\pm$.007} & {\tiny$\pm$.013} & {\tiny$\pm$.013} & {\tiny$\pm$.005} & {\tiny$\pm$.003} & {\tiny$\pm$.006} & {\tiny$\pm$.006} & {\tiny$\pm$.002} & {\tiny$\pm$.008} & {\tiny$\pm$.003} & {\tiny$\pm$.004} & {\tiny$\pm$.004} & {\tiny$\pm$.010} & {\tiny$\pm$.001} & {\tiny$\pm$.004} & {\tiny$\pm$.008} & {\tiny$\pm$.006} & {\tiny$\pm$.010} & {\tiny$\pm$.007} \\[1pt]
deeploc\_binary/default & \textbf{0.908} & \textbf{0.908} & 0.906 & 0.902 & 0.901 & 0.902 & 0.900 & 0.906 & \textbf{0.908} & 0.894 & 0.897 & 0.907 & 0.905 & 0.901 & 0.906 & 0.905 & 0.905 & 0.904 & 0.900 & 0.893 & 0.894 & 0.896 \\
 & {\tiny$\pm$.004} & {\tiny$\pm$.004} & {\tiny$\pm$.002} & {\tiny$\pm$.006} & {\tiny$\pm$.004} & {\tiny$\pm$.001} & {\tiny$\pm$.003} & {\tiny$\pm$.004} & {\tiny$\pm$.005} & {\tiny$\pm$.003} & {\tiny$\pm$.007} & {\tiny$\pm$.007} & {\tiny$\pm$.003} & {\tiny$\pm$.006} & {\tiny$\pm$.002} & {\tiny$\pm$.002} & {\tiny$\pm$.004} & {\tiny$\pm$.003} & {\tiny$\pm$.003} & {\tiny$\pm$.004} & {\tiny$\pm$.008} & {\tiny$\pm$.006} \\[1pt]
deeploc\_2/test & \textbf{0.930} & 0.924 & 0.925 & 0.922 & 0.922 & 0.915 & 0.923 & 0.920 & 0.925 & 0.925 & 0.920 & 0.915 & 0.922 & 0.916 & 0.918 & 0.922 & 0.923 & 0.923 & 0.920 & 0.919 & 0.917 & 0.918 \\
 & {\tiny$\pm$.001} & {\tiny$\pm$.000} & {\tiny$\pm$.001} & {\tiny$\pm$.001} & {\tiny$\pm$.001} & {\tiny$\pm$.003} & {\tiny$\pm$.002} & {\tiny$\pm$.001} & {\tiny$\pm$.003} & {\tiny$\pm$.002} & {\tiny$\pm$.002} & {\tiny$\pm$.002} & {\tiny$\pm$.002} & {\tiny$\pm$.001} & {\tiny$\pm$.001} & {\tiny$\pm$.002} & {\tiny$\pm$.001} & {\tiny$\pm$.000} & {\tiny$\pm$.003} & {\tiny$\pm$.002} & {\tiny$\pm$.002} & {\tiny$\pm$.002} \\[1pt]
remote\_homology/family\_holdout & \textbf{0.829} & 0.704 & 0.672 & 0.674 & 0.648 & 0.650 & 0.637 & 0.621 & 0.708 & 0.696 & 0.659 & 0.628 & 0.684 & 0.637 & 0.660 & 0.672 & 0.683 & 0.620 & 0.628 & 0.590 & 0.636 & 0.616 \\
 & {\tiny$\pm$.014} & {\tiny$\pm$.024} & {\tiny$\pm$.023} & {\tiny$\pm$.022} & {\tiny$\pm$.012} & {\tiny$\pm$.019} & {\tiny$\pm$.044} & {\tiny$\pm$.041} & {\tiny$\pm$.016} & {\tiny$\pm$.022} & {\tiny$\pm$.046} & {\tiny$\pm$.035} & {\tiny$\pm$.008} & {\tiny$\pm$.039} & {\tiny$\pm$.030} & {\tiny$\pm$.024} & {\tiny$\pm$.001} & {\tiny$\pm$.027} & {\tiny$\pm$.015} & {\tiny$\pm$.047} & {\tiny$\pm$.022} & {\tiny$\pm$.010} \\[1pt]
remote\_homology/superfamily\_holdout & \textbf{0.411} & 0.273 & 0.248 & 0.236 & 0.221 & 0.227 & 0.219 & 0.219 & 0.268 & 0.236 & 0.249 & 0.218 & 0.250 & 0.208 & 0.213 & 0.258 & 0.256 & 0.232 & 0.225 & 0.200 & 0.201 & 0.193 \\
 & {\tiny$\pm$.006} & {\tiny$\pm$.003} & {\tiny$\pm$.007} & {\tiny$\pm$.016} & {\tiny$\pm$.005} & {\tiny$\pm$.004} & {\tiny$\pm$.012} & {\tiny$\pm$.021} & {\tiny$\pm$.003} & {\tiny$\pm$.006} & {\tiny$\pm$.010} & {\tiny$\pm$.009} & {\tiny$\pm$.002} & {\tiny$\pm$.009} & {\tiny$\pm$.009} & {\tiny$\pm$.006} & {\tiny$\pm$.007} & {\tiny$\pm$.010} & {\tiny$\pm$.007} & {\tiny$\pm$.009} & {\tiny$\pm$.007} & {\tiny$\pm$.007} \\[1pt]
remote\_homology/fold\_holdout & \textbf{0.248} & 0.215 & 0.213 & 0.206 & 0.195 & 0.189 & 0.193 & 0.196 & 0.223 & 0.205 & 0.222 & 0.215 & 0.201 & 0.181 & 0.190 & 0.197 & 0.199 & 0.188 & 0.192 & 0.179 & 0.179 & 0.170 \\
 & {\tiny$\pm$.007} & {\tiny$\pm$.003} & {\tiny$\pm$.008} & {\tiny$\pm$.011} & {\tiny$\pm$.002} & {\tiny$\pm$.013} & {\tiny$\pm$.013} & {\tiny$\pm$.014} & {\tiny$\pm$.003} & {\tiny$\pm$.006} & {\tiny$\pm$.006} & {\tiny$\pm$.011} & {\tiny$\pm$.005} & {\tiny$\pm$.011} & {\tiny$\pm$.011} & {\tiny$\pm$.010} & {\tiny$\pm$.007} & {\tiny$\pm$.002} & {\tiny$\pm$.013} & {\tiny$\pm$.007} & {\tiny$\pm$.010} & {\tiny$\pm$.005} \\[1pt]
ppi\_shs27k/default & 0.534 & 0.530 & 0.529 & 0.526 & 0.534 & 0.535 & 0.536 & 0.534 & 0.527 & \textbf{0.538} & 0.534 & 0.529 & 0.534 & 0.535 & 0.537 & 0.526 & 0.531 & 0.526 & 0.522 & 0.525 & 0.535 & 0.530 \\
 & {\tiny$\pm$.005} & {\tiny$\pm$.006} & {\tiny$\pm$.011} & {\tiny$\pm$.005} & {\tiny$\pm$.004} & {\tiny$\pm$.008} & {\tiny$\pm$.004} & {\tiny$\pm$.004} & {\tiny$\pm$.005} & {\tiny$\pm$.006} & {\tiny$\pm$.009} & {\tiny$\pm$.011} & {\tiny$\pm$.009} & {\tiny$\pm$.006} & {\tiny$\pm$.006} & {\tiny$\pm$.004} & {\tiny$\pm$.009} & {\tiny$\pm$.004} & {\tiny$\pm$.005} & {\tiny$\pm$.002} & {\tiny$\pm$.004} & {\tiny$\pm$.005} \\[1pt]
\bottomrule
\end{tabular}
\end{sidewaystable}

\begin{sidewaystable}[p]\centering
\caption{\textbf{Scratch} --- single-layer transformer encoder (width 320, 8 heads) trained on each task without pretraining; every task/split. PUMA and PUMA\_PC are the BLOSUM62,\,0.7,\,0.05 configuration. Mean over seeds with the seed standard deviation beneath; higher is better; 
\textbf{bold} marks the best value in each row, including ties. Without a pretrained encoder these numbers isolate what the tokenizer alone contributes. The sign of the difference from AA changes between task families, and it is the same for BPE, PUMA and PUMA\_PC. The median seed standard deviation is 0.015, and many exceed 0.05 (up to 0.24), comparable to many of the differences between cells.}
\label{sup:tab:sc}
\tiny\setlength{\tabcolsep}{3pt}
\begin{tabular}{l c ccccccc ccccccc ccccccc}
\toprule
& AA & \multicolumn{7}{c}{BPE} & \multicolumn{7}{c}{PUMA} & \multicolumn{7}{c}{PUMA\_PC}\\
\cmidrule(lr){3-9}\cmidrule(lr){10-16}\cmidrule(lr){17-23}
Task/split & 33 & .8k & 1.6k & 3.2k & 6.4k & 12.8k & 25.6k & 51.2k & .8k & 1.6k & 3.2k & 6.4k & 12.8k & 25.6k & 51.2k & .8k & 1.6k & 3.2k & 6.4k & 12.8k & 25.6k & 51.2k\\
\midrule
fluorescence/default & 0.305 & 0.569 & 0.573 & 0.585 & 0.592 & 0.599 & 0.636 & \textbf{0.650} & 0.560 & 0.510 & 0.561 & 0.627 & 0.634 & 0.614 & 0.625 & 0.522 & 0.588 & 0.536 & 0.616 & 0.614 & 0.504 & 0.618 \\
 & {\tiny$\pm$.142} & {\tiny$\pm$.024} & {\tiny$\pm$.042} & {\tiny$\pm$.020} & {\tiny$\pm$.028} & {\tiny$\pm$.024} & {\tiny$\pm$.017} & {\tiny$\pm$.014} & {\tiny$\pm$.007} & {\tiny$\pm$.110} & {\tiny$\pm$.051} & {\tiny$\pm$.043} & {\tiny$\pm$.018} & {\tiny$\pm$.021} & {\tiny$\pm$.018} & {\tiny$\pm$.030} & {\tiny$\pm$.012} & {\tiny$\pm$.042} & {\tiny$\pm$.024} & {\tiny$\pm$.019} & {\tiny$\pm$.149} & {\tiny$\pm$.009} \\[1pt]
stability/default & \textbf{0.641} & 0.523 & 0.551 & 0.623 & 0.582 & 0.365 & 0.500 & 0.496 & 0.539 & 0.577 & 0.576 & 0.555 & 0.499 & 0.479 & 0.456 & 0.550 & 0.550 & 0.522 & 0.454 & 0.374 & 0.381 & 0.469 \\
 & {\tiny$\pm$.039} & {\tiny$\pm$.071} & {\tiny$\pm$.006} & {\tiny$\pm$.023} & {\tiny$\pm$.048} & {\tiny$\pm$.240} & {\tiny$\pm$.015} & {\tiny$\pm$.018} & {\tiny$\pm$.019} & {\tiny$\pm$.039} & {\tiny$\pm$.014} & {\tiny$\pm$.022} & {\tiny$\pm$.008} & {\tiny$\pm$.027} & {\tiny$\pm$.092} & {\tiny$\pm$.013} & {\tiny$\pm$.034} & {\tiny$\pm$.039} & {\tiny$\pm$.047} & {\tiny$\pm$.038} & {\tiny$\pm$.071} & {\tiny$\pm$.018} \\[1pt]
aav/two\_vs\_many & 0.326 & 0.343 & 0.232 & 0.221 & 0.448 & 0.446 & 0.472 & 0.422 & 0.086 & 0.483 & 0.206 & 0.325 & \textbf{0.578} & 0.492 & 0.524 & 0.234 & 0.454 & 0.328 & 0.368 & 0.486 & 0.340 & 0.576 \\
 & {\tiny$\pm$.008} & {\tiny$\pm$.061} & {\tiny$\pm$.210} & {\tiny$\pm$.166} & {\tiny$\pm$.037} & {\tiny$\pm$.167} & {\tiny$\pm$.124} & {\tiny$\pm$.041} & {\tiny$\pm$.138} & {\tiny$\pm$.028} & {\tiny$\pm$.072} & {\tiny$\pm$.089} & {\tiny$\pm$.054} & {\tiny$\pm$.051} & {\tiny$\pm$.037} & {\tiny$\pm$.165} & {\tiny$\pm$.064} & {\tiny$\pm$.072} & {\tiny$\pm$.070} & {\tiny$\pm$.109} & {\tiny$\pm$.140} & {\tiny$\pm$.027} \\[1pt]
gb1/two\_vs\_rest & 0.286 & 0.383 & 0.434 & 0.181 & 0.428 & 0.161 & 0.130 & 0.403 & 0.284 & \textbf{0.470} & 0.267 & 0.232 & 0.150 & 0.081 & 0.437 & 0.367 & 0.350 & 0.232 & 0.218 & 0.048 & 0.081 & 0.368 \\
 & {\tiny$\pm$.019} & {\tiny$\pm$.104} & {\tiny$\pm$.021} & {\tiny$\pm$.123} & {\tiny$\pm$.014} & {\tiny$\pm$.163} & {\tiny$\pm$.125} & {\tiny$\pm$.017} & {\tiny$\pm$.064} & {\tiny$\pm$.069} & {\tiny$\pm$.082} & {\tiny$\pm$.023} & {\tiny$\pm$.117} & {\tiny$\pm$.075} & {\tiny$\pm$.096} & {\tiny$\pm$.012} & {\tiny$\pm$.104} & {\tiny$\pm$.074} & {\tiny$\pm$.134} & {\tiny$\pm$.033} & {\tiny$\pm$.075} & {\tiny$\pm$.108} \\[1pt]
gb1/three\_vs\_rest & \textbf{0.860} & 0.768 & 0.782 & 0.789 & 0.777 & 0.772 & 0.785 & 0.787 & 0.763 & 0.744 & 0.767 & 0.748 & 0.742 & 0.743 & 0.748 & 0.765 & 0.759 & 0.769 & 0.757 & 0.756 & 0.753 & 0.771 \\
 & {\tiny$\pm$.004} & {\tiny$\pm$.004} & {\tiny$\pm$.001} & {\tiny$\pm$.002} & {\tiny$\pm$.008} & {\tiny$\pm$.002} & {\tiny$\pm$.003} & {\tiny$\pm$.013} & {\tiny$\pm$.015} & {\tiny$\pm$.003} & {\tiny$\pm$.006} & {\tiny$\pm$.008} & {\tiny$\pm$.009} & {\tiny$\pm$.010} & {\tiny$\pm$.004} & {\tiny$\pm$.012} & {\tiny$\pm$.002} & {\tiny$\pm$.007} & {\tiny$\pm$.002} & {\tiny$\pm$.004} & {\tiny$\pm$.002} & {\tiny$\pm$.004} \\[1pt]
gb1/low\_vs\_high & \textbf{0.361} & 0.277 & 0.244 & 0.232 & 0.231 & 0.245 & 0.239 & 0.269 & 0.247 & 0.183 & 0.231 & 0.219 & 0.211 & 0.230 & 0.246 & 0.222 & 0.179 & 0.226 & 0.197 & 0.231 & 0.246 & 0.241 \\
 & {\tiny$\pm$.047} & {\tiny$\pm$.032} & {\tiny$\pm$.019} & {\tiny$\pm$.018} & {\tiny$\pm$.031} & {\tiny$\pm$.033} & {\tiny$\pm$.026} & {\tiny$\pm$.026} & {\tiny$\pm$.022} & {\tiny$\pm$.059} & {\tiny$\pm$.031} & {\tiny$\pm$.032} & {\tiny$\pm$.009} & {\tiny$\pm$.018} & {\tiny$\pm$.006} & {\tiny$\pm$.025} & {\tiny$\pm$.029} & {\tiny$\pm$.011} & {\tiny$\pm$.050} & {\tiny$\pm$.023} & {\tiny$\pm$.026} & {\tiny$\pm$.015} \\[1pt]
solmut\_cs/default & 0.041 & 0.150 & 0.139 & 0.170 & 0.119 & 0.141 & 0.220 & 0.226 & 0.083 & 0.160 & 0.137 & 0.202 & 0.221 & 0.243 & \textbf{0.263} & 0.125 & 0.206 & 0.174 & 0.232 & 0.151 & 0.207 & 0.223 \\
 & {\tiny$\pm$.080} & {\tiny$\pm$.073} & {\tiny$\pm$.013} & {\tiny$\pm$.050} & {\tiny$\pm$.113} & {\tiny$\pm$.028} & {\tiny$\pm$.035} & {\tiny$\pm$.016} & {\tiny$\pm$.012} & {\tiny$\pm$.087} & {\tiny$\pm$.009} & {\tiny$\pm$.022} & {\tiny$\pm$.077} & {\tiny$\pm$.042} & {\tiny$\pm$.047} & {\tiny$\pm$.043} & {\tiny$\pm$.067} & {\tiny$\pm$.036} & {\tiny$\pm$.034} & {\tiny$\pm$.143} & {\tiny$\pm$.022} & {\tiny$\pm$.037} \\[1pt]
solmut\_blat/default & 0.093 & 0.123 & 0.159 & 0.152 & 0.160 & 0.166 & 0.143 & 0.170 & 0.113 & 0.126 & 0.125 & 0.160 & 0.118 & 0.202 & \textbf{0.233} & 0.093 & 0.135 & 0.109 & 0.165 & 0.171 & 0.119 & 0.167 \\
 & {\tiny$\pm$.039} & {\tiny$\pm$.016} & {\tiny$\pm$.054} & {\tiny$\pm$.034} & {\tiny$\pm$.052} & {\tiny$\pm$.011} & {\tiny$\pm$.066} & {\tiny$\pm$.023} & {\tiny$\pm$.018} & {\tiny$\pm$.032} & {\tiny$\pm$.016} & {\tiny$\pm$.059} & {\tiny$\pm$.031} & {\tiny$\pm$.044} & {\tiny$\pm$.038} & {\tiny$\pm$.030} & {\tiny$\pm$.026} & {\tiny$\pm$.020} & {\tiny$\pm$.038} & {\tiny$\pm$.062} & {\tiny$\pm$.034} & {\tiny$\pm$.007} \\[1pt]
solmut\_lgk/default & -0.006 & 0.055 & 0.075 & 0.104 & 0.135 & \textbf{0.162} & 0.126 & 0.080 & 0.119 & 0.065 & 0.124 & 0.149 & 0.115 & 0.143 & 0.094 & 0.114 & 0.075 & 0.077 & 0.121 & 0.094 & 0.070 & 0.121 \\
 & {\tiny$\pm$.037} & {\tiny$\pm$.024} & {\tiny$\pm$.050} & {\tiny$\pm$.042} & {\tiny$\pm$.038} & {\tiny$\pm$.019} & {\tiny$\pm$.039} & {\tiny$\pm$.024} & {\tiny$\pm$.051} & {\tiny$\pm$.011} & {\tiny$\pm$.044} & {\tiny$\pm$.045} & {\tiny$\pm$.052} & {\tiny$\pm$.045} & {\tiny$\pm$.066} & {\tiny$\pm$.014} & {\tiny$\pm$.005} & {\tiny$\pm$.027} & {\tiny$\pm$.013} & {\tiny$\pm$.068} & {\tiny$\pm$.047} & {\tiny$\pm$.137} \\[1pt]
meltome/mixed\_split & \textbf{0.499} & 0.458 & 0.433 & 0.429 & 0.420 & 0.387 & 0.353 & 0.354 & 0.451 & 0.446 & 0.422 & 0.411 & 0.390 & 0.366 & 0.350 & 0.448 & 0.444 & 0.433 & 0.421 & 0.395 & 0.383 & 0.364 \\
 & {\tiny$\pm$.009} & {\tiny$\pm$.001} & {\tiny$\pm$.006} & {\tiny$\pm$.008} & {\tiny$\pm$.004} & {\tiny$\pm$.005} & {\tiny$\pm$.012} & {\tiny$\pm$.006} & {\tiny$\pm$.006} & {\tiny$\pm$.003} & {\tiny$\pm$.008} & {\tiny$\pm$.016} & {\tiny$\pm$.010} & {\tiny$\pm$.022} & {\tiny$\pm$.007} & {\tiny$\pm$.012} & {\tiny$\pm$.011} & {\tiny$\pm$.013} & {\tiny$\pm$.012} & {\tiny$\pm$.005} & {\tiny$\pm$.009} & {\tiny$\pm$.011} \\[1pt]
meltome/human & \textbf{0.546} & 0.495 & 0.469 & 0.485 & 0.463 & 0.448 & 0.446 & 0.433 & 0.497 & 0.494 & 0.465 & 0.419 & 0.434 & 0.439 & 0.427 & 0.498 & 0.499 & 0.489 & 0.485 & 0.487 & 0.467 & 0.481 \\
 & {\tiny$\pm$.013} & {\tiny$\pm$.010} & {\tiny$\pm$.022} & {\tiny$\pm$.008} & {\tiny$\pm$.008} & {\tiny$\pm$.019} & {\tiny$\pm$.021} & {\tiny$\pm$.017} & {\tiny$\pm$.030} & {\tiny$\pm$.012} & {\tiny$\pm$.023} & {\tiny$\pm$.044} & {\tiny$\pm$.004} & {\tiny$\pm$.006} & {\tiny$\pm$.022} & {\tiny$\pm$.011} & {\tiny$\pm$.013} & {\tiny$\pm$.012} & {\tiny$\pm$.012} & {\tiny$\pm$.008} & {\tiny$\pm$.010} & {\tiny$\pm$.001} \\[1pt]
deepsol/default & 0.513 & 0.637 & 0.592 & 0.593 & 0.559 & 0.566 & 0.612 & 0.627 & 0.568 & 0.631 & 0.637 & 0.623 & \textbf{0.644} & 0.623 & 0.633 & 0.571 & 0.569 & 0.531 & 0.594 & 0.604 & 0.632 & 0.642 \\
 & {\tiny$\pm$.016} & {\tiny$\pm$.003} & {\tiny$\pm$.053} & {\tiny$\pm$.054} & {\tiny$\pm$.059} & {\tiny$\pm$.065} & {\tiny$\pm$.006} & {\tiny$\pm$.004} & {\tiny$\pm$.064} & {\tiny$\pm$.015} & {\tiny$\pm$.006} & {\tiny$\pm$.015} & {\tiny$\pm$.021} & {\tiny$\pm$.007} & {\tiny$\pm$.009} & {\tiny$\pm$.070} & {\tiny$\pm$.069} & {\tiny$\pm$.053} & {\tiny$\pm$.054} & {\tiny$\pm$.060} & {\tiny$\pm$.005} & {\tiny$\pm$.009} \\[1pt]
deeploc\_binary/default & 0.749 & 0.735 & 0.727 & 0.713 & 0.712 & 0.721 & 0.713 & 0.702 & 0.725 & 0.725 & 0.714 & 0.701 & 0.709 & 0.706 & 0.669 & 0.731 & \textbf{0.756} & 0.735 & 0.742 & 0.738 & 0.731 & 0.717 \\
 & {\tiny$\pm$.002} & {\tiny$\pm$.008} & {\tiny$\pm$.008} & {\tiny$\pm$.018} & {\tiny$\pm$.012} & {\tiny$\pm$.011} & {\tiny$\pm$.007} & {\tiny$\pm$.003} & {\tiny$\pm$.015} & {\tiny$\pm$.017} & {\tiny$\pm$.015} & {\tiny$\pm$.014} & {\tiny$\pm$.005} & {\tiny$\pm$.014} & {\tiny$\pm$.017} & {\tiny$\pm$.009} & {\tiny$\pm$.007} & {\tiny$\pm$.006} & {\tiny$\pm$.006} & {\tiny$\pm$.015} & {\tiny$\pm$.017} & {\tiny$\pm$.011} \\[1pt]
deeploc\_2/test & \textbf{0.902} & 0.894 & 0.892 & 0.888 & 0.886 & 0.883 & 0.881 & 0.881 & 0.894 & 0.893 & 0.890 & 0.886 & 0.882 & 0.883 & 0.879 & 0.897 & 0.894 & 0.891 & 0.891 & 0.891 & 0.890 & 0.889 \\
 & {\tiny$\pm$.001} & {\tiny$\pm$.000} & {\tiny$\pm$.001} & {\tiny$\pm$.001} & {\tiny$\pm$.002} & {\tiny$\pm$.002} & {\tiny$\pm$.002} & {\tiny$\pm$.002} & {\tiny$\pm$.000} & {\tiny$\pm$.001} & {\tiny$\pm$.001} & {\tiny$\pm$.001} & {\tiny$\pm$.003} & {\tiny$\pm$.002} & {\tiny$\pm$.001} & {\tiny$\pm$.001} & {\tiny$\pm$.001} & {\tiny$\pm$.001} & {\tiny$\pm$.001} & {\tiny$\pm$.003} & {\tiny$\pm$.001} & {\tiny$\pm$.002} \\[1pt]
remote\_homology/family\_holdout & 0.283 & 0.306 & 0.329 & 0.261 & 0.249 & 0.209 & 0.236 & 0.206 & 0.332 & 0.285 & 0.295 & 0.298 & 0.254 & 0.229 & 0.218 & \textbf{0.351} & 0.314 & 0.244 & 0.270 & 0.308 & 0.305 & 0.186 \\
 & {\tiny$\pm$.006} & {\tiny$\pm$.032} & {\tiny$\pm$.063} & {\tiny$\pm$.013} & {\tiny$\pm$.045} & {\tiny$\pm$.015} & {\tiny$\pm$.009} & {\tiny$\pm$.019} & {\tiny$\pm$.045} & {\tiny$\pm$.032} & {\tiny$\pm$.020} & {\tiny$\pm$.028} & {\tiny$\pm$.010} & {\tiny$\pm$.015} & {\tiny$\pm$.013} & {\tiny$\pm$.043} & {\tiny$\pm$.014} & {\tiny$\pm$.014} & {\tiny$\pm$.028} & {\tiny$\pm$.051} & {\tiny$\pm$.036} & {\tiny$\pm$.050} \\[1pt]
remote\_homology/superfamily\_holdout & 0.074 & 0.077 & 0.077 & 0.074 & 0.065 & 0.060 & 0.057 & 0.061 & 0.078 & 0.072 & 0.065 & 0.069 & 0.058 & 0.065 & 0.055 & \textbf{0.085} & 0.078 & 0.077 & 0.064 & 0.069 & 0.065 & 0.058 \\
 & {\tiny$\pm$.007} & {\tiny$\pm$.005} & {\tiny$\pm$.004} & {\tiny$\pm$.004} & {\tiny$\pm$.001} & {\tiny$\pm$.005} & {\tiny$\pm$.002} & {\tiny$\pm$.003} & {\tiny$\pm$.002} & {\tiny$\pm$.005} & {\tiny$\pm$.005} & {\tiny$\pm$.004} & {\tiny$\pm$.003} & {\tiny$\pm$.006} & {\tiny$\pm$.002} & {\tiny$\pm$.003} & {\tiny$\pm$.005} & {\tiny$\pm$.007} & {\tiny$\pm$.005} & {\tiny$\pm$.000} & {\tiny$\pm$.005} & {\tiny$\pm$.002} \\[1pt]
remote\_homology/fold\_holdout & 0.069 & 0.092 & 0.089 & 0.099 & 0.088 & 0.088 & 0.092 & 0.085 & 0.100 & 0.103 & \textbf{0.108} & 0.090 & 0.087 & 0.092 & 0.086 & 0.093 & 0.084 & 0.086 & 0.090 & 0.074 & 0.086 & 0.095 \\
 & {\tiny$\pm$.038} & {\tiny$\pm$.010} & {\tiny$\pm$.012} & {\tiny$\pm$.007} & {\tiny$\pm$.009} & {\tiny$\pm$.005} & {\tiny$\pm$.008} & {\tiny$\pm$.007} & {\tiny$\pm$.017} & {\tiny$\pm$.002} & {\tiny$\pm$.006} & {\tiny$\pm$.004} & {\tiny$\pm$.007} & {\tiny$\pm$.007} & {\tiny$\pm$.006} & {\tiny$\pm$.007} & {\tiny$\pm$.009} & {\tiny$\pm$.015} & {\tiny$\pm$.005} & {\tiny$\pm$.006} & {\tiny$\pm$.005} & {\tiny$\pm$.001} \\[1pt]
ppi\_shs27k/default & 0.548 & 0.556 & 0.563 & 0.558 & 0.568 & \textbf{0.570} & 0.554 & 0.569 & 0.557 & 0.559 & 0.564 & 0.559 & 0.562 & 0.568 & 0.561 & 0.554 & 0.556 & 0.558 & 0.559 & 0.552 & 0.556 & 0.564 \\
 & {\tiny$\pm$.008} & {\tiny$\pm$.004} & {\tiny$\pm$.011} & {\tiny$\pm$.002} & {\tiny$\pm$.004} & {\tiny$\pm$.004} & {\tiny$\pm$.006} & {\tiny$\pm$.001} & {\tiny$\pm$.002} & {\tiny$\pm$.005} & {\tiny$\pm$.007} & {\tiny$\pm$.006} & {\tiny$\pm$.005} & {\tiny$\pm$.004} & {\tiny$\pm$.002} & {\tiny$\pm$.004} & {\tiny$\pm$.003} & {\tiny$\pm$.007} & {\tiny$\pm$.008} & {\tiny$\pm$.006} & {\tiny$\pm$.009} & {\tiny$\pm$.003} \\[1pt]
\bottomrule
\end{tabular}
\end{sidewaystable}

\end{document}